\documentclass[runningheads]{llncs}
\pdfoutput=1
\usepackage{eccv}

\usepackage{eccvabbrv}

\usepackage{graphicx}
\usepackage{float}
\usepackage{booktabs}

\usepackage[accsupp]{axessibility}  %

\usepackage{array}
\usepackage{fontawesome5}
\usepackage{float}
\usepackage{multirow}
\usepackage{arydshln} %
\usepackage[dvipsnames]{xcolor}

\usepackage{pgf}
\usepackage{pgfplots}

\everymath=\expandafter{\the\everymath\displaystyle}
\makeatletter\@ifpackageloaded{underscore}{}{\usepackage[strings]{underscore}}

\makeatother
\usepackage{tikz}
\usepackage{pgfplots}
\pgfplotsset{compat=1.14}

\usetikzlibrary{shapes}
\usetikzlibrary{calc}
\usetikzlibrary{backgrounds}
\usetikzlibrary{positioning}
\usetikzlibrary{arrows}
\usetikzlibrary{arrows.meta}
\usetikzlibrary{decorations.text}    
\usetikzlibrary{fit}
\usetikzlibrary{spy}
\usetikzlibrary{3d}
\usetikzlibrary{positioning}
\usepackage[weather]{ifsym}

\usepackage[skins]{tcolorbox}

\definecolor{turquoise}{cmyk}{0.65,0,0.1,0.3}
\definecolor{purple}{rgb}{0.65,0,0.65}
\definecolor{dark_green}{rgb}{0, 0.5, 0}
\definecolor{light-green}{rgb}{0.2, 0.7, 0.2}
\definecolor{orange}{rgb}{0.7, 0.5, 0.5}
\definecolor{orangeish}{rgb}{0.8, 0.7, 0.1}
\definecolor{ipadaptercolor}{rgb}{0.8, 0.7, 0.1}
\definecolor{red}{rgb}{0.8, 0.2, 0.2}
\definecolor{darkred}{rgb}{0.6, 0.1, 0.05}
\definecolor{blueish}{rgb}{0.0, 0.3, .6}
\definecolor{light_gray}{rgb}{0.7, 0.7, .7}
\definecolor{dark-gray}{rgb}{0.3, 0.3, .3}
\definecolor{pink}{rgb}{1, 0, 1}
\definecolor{greyblue}{rgb}{0.25, 0.25, 1}

\usepackage{hyperref}

\usepackage{orcidlink}

\title{IPAdapter-Instruct: Resolving Ambiguity in Image-based Conditioning using Instruct Prompts}
\titlerunning{IPAdapter-Instruct}

\author{Ciara Rowles, Shimon Vainer, Dante De Nigris, Slava Elizarov, Konstantin Kutsy, Simon Donn\'{e}}

\authorrunning{C.~Rowles et al.}

{\tt\small }
\institute{Unity Technologies\\
Corresponding author: \href{mailto:ciara.rowles@unity3d.com}{ciara.rowles@unity3d.com}
}

\usepackage[width=122mm,left=12mm,paperwidth=146mm,height=193mm,top=12mm,paperheight=217mm]{geometry}
\begin{document}
\maketitle
\begin{figure}%
    \input{fig/teaser_new.tex}
\end{figure}

\begin{abstract}%
Diffusion models continuously push the boundary of state-of-the-art image generation, but the process is hard to control with any nuance: practice proves that textual prompts are inadequate for accurately describing image style or fine structural details (such as faces).
ControlNet~\cite{ControlNet} and IPAdapter~\cite{IPAdapter} address this shortcoming by conditioning the generative process on imagery instead, but each individual instance is limited to modeling a single conditional posterior: for practical use-cases, where multiple different posteriors are desired within the same workflow, training and using multiple adapters is cumbersome.
We propose IPAdapter-Instruct, which combines natural-image conditioning with ``Instruct'' prompts to swap between interpretations for the same conditioning image: style transfer, object extraction, both, or something else still?
IPAdapterInstruct efficiently learns multiple tasks with minimal loss in quality compared to dedicated per-task models.

\end{abstract}

\section{Introduction}\label{sec:intro}%

The field of image generation has received fresh impetus from diffusion model theory.
Modeling image generation as an iterative process that reverses the diffusion of images into pure noise, diffusion models have achieved state-of-the-art performance in image generation tasks.
They have proven to be more robust to train than the previous GAN-based state-of-the-art models, and to be more widely applicable in practice: a key aspect of which is the ease of class-free conditioning of the generation on text prompts using classifier-free guidance, offering unprecedented control over the output image.

However, the common proverb ``An image is worth a thousand words'' is as relevant as ever: crafting text prompts that result in exactly the desired image is a non-trivial task, referred to as \emph{prompt engineering}.
Even when the desired image is clear in the mind of the user, it is difficult to express it in such a way that the model output replicates it accurately; expressing intent through images is often far more intuitive than through text.
This concept has spurred the development of image-based conditioning such as ControlNet\cite{ControlNet} and IPAdapter\cite{IPAdapter} approaches.
From scribbles and sketches or stylistic examples to many other modalities: these methods allow users to express their intent in the image domain, where spatial information and stylistic cues can be more easily conveyed.
Yet in practice every instance of these methods only offers a single way of guiding the generation based on the image prompt: when a user wishes to alternate between different ways of conditioning or combine them, each of the models needs to be trained separately.
Managing the various models to correctly combine them and switch between them is complex and cumbersome.

We aim to tackle the ambiguity of conditioning with images.
In the case of ControlNets, which condition the image generation on \eg{} canny edge maps, or depth images, or normal images, the intention of the user is unambiguous: ``generate an image such that <modality> extracted from it matches my condition image''.
However, when conditioning with natural RGB images, the intention is unclear: is the user asking for generated images that have similar style? That contain the same object? That are set in the same background? That have the same composition?
We propose \textbf{IPAdapter-Instruct} as a solution: a single model that can be textually instructed how to interpret the condition image, a task for which text instructions are well suited.
Training such a model to handle multiple conditioning aspects is not only feasible, but even results in a flexible model that is on-par with task-specific models, as we show in our ablation experiments.

We discuss how to condition on this text instruction, as well as train a specific instance with five ways of interpreting the condition image: total replication (as the original IPAdapter~\cite{IPAdapter}), transferring the style, replicating the composition, placing an object in a new image, and placing a person in a new image.

\section{Related Work}\label{sec:related}

\subsection{Diffusion Models}

Diffusion models generate images conditioned on text by learning to reverse a gradual diffusion process~\cite{DiffusionTheory,DDPM,DDIM}, typically in a latent-pixel domain for its efficiency and low-level prior~\cite{LatentDiffusionModels,SD3}.
Unfortunately, these diffusion models are costly to train: because of memory and computational requirements, and because of the need for immense volumes of training data~\cite{LAION5B}.
While much work is focused on inference speed or distillation into smaller models, we consider those efforts orthogonal to our work and omit the relevant literature for brevity's sake.

However, the text prompts that condition these models are finicky and inaccurate for conveying user intent~\cite{PromptEngineering}, with some authors instead ``translating'' user prompts into model-aligned ``system language''~\cite{PromptRefinementImagePivot,AutomaticPromptRefinement}.
Although negative prompts provide additional control, they can interfere with the original prompt or even be ignored~\cite{NegativePrompts}, while still  being restricted in their expressiveness.
These difficulties imply a need for more expressive control, which we believe to be in images.
Therefore, we now discuss both image-based conditioning for diffusion models and diffusion models for image-to-image translation tasks.

\subsection{Image-based Control for Diffusion Models}

\textbf{IPAdapter}~\cite{IPAdapter} comprises a small neural translator to project from the input image's embedding, such as from ViT-H/14~\cite{ViT-H/14CLIP} CLIP~\cite{OpenAI-CLIP}, onto the embedding space used by the text encoder; the network cross-attends to these novel embeddings in additional cross-attention layers similar to those of the text prompt, effectively enabling it to use an image as prompt input.
IPAdapter is trained to reproduce the input image exactly --- only ``by coincidence'' is the emergent behavior of the model to flexibly transfer the style and content of the condition image to the output images.
Our IPAdapter-Instruct approach both makes the desired behavior explicit (through an ``Instruct'' prompt) and supervises the various behaviors directly during training (through task-specific datasets).

\textbf{UnCLIP-based Approaches}~\cite{UnCLIP} retrain the base model to reproduce an image based on its CLIP embedding, similar to IPAdapter, but replace the text prompt with the image condition completely, attaining a single mode of control over the output.
Since the entire model is retrained for this purpose, it risks catastrophic forgetting of the original model's capabilities, and is incompatible with other residual changes to the base model such as LoRA's~\cite{LORA}: we instead prefer to build off IPAdapter for keeping the base model intact.

\textbf{Composer Models}~\cite{Composer} retrain the diffusion model by conditioning it on a large set of control signals automatically extracted from the target image that intend to exhaustively describe it, with the individual conditioning modalities assumed to be unambiguous in their intention.
While this method offers the highest level of control, the entire model must be retrained to achieve this (similar to unCLIP), resulting in no compatibility with other models (such as LoRAs) and no straightforward way to retrain an existing model.

\textbf{InstantStyle}~\cite{InstantStyle} conditions the output on the style of an input image without training.
To do so, the CLIP space embedding of its textual description is subtracted from that of the image to obtain a ``style direction vector'' in CLIP space: the text prompt cross attention layers in some blocks are then extended to also attends to this style vector.
While this model is neither explicitly trained to model the conditioning process, its success is undeniable.
We integrate this task (by generating a task dataset using InstantStyle) into the training process for only a small overhead, as part of the general training procedure.

\textbf{ControlNet}~\cite{ControlNet} functions by cloning the base model's encoder.
The clone's inputs are replaced with the conditioning image, and its outputs are added as residuals to the original model's hidden states~\cite{ControlNet,ControlNet-XS}. 
Compared to IPAdapter, ControlNet input modalities tend to vary more widely: canny edge responses, normal maps, densely encoded poses, \etc.
While ControlNet models are mostly task-specific, but ControlNet++~\cite{ControlNet++} flexibly handles multiple input modalities: but as there is little ambiguity between the  various conditions, they found no need for an ``Instruct'' prompt.
We find that these models implement much more spatially localized control over the output compared to IPAdapter, while being more flexible and interoperable than the image-to-image models discussed below: ControlNets are compatible with IPAdapters, as they both keep the base model intact and only perform residual edits on the base model's hidden states.

\subsection{Image-to-Image Diffusion Models and Instruct-based Control}

Image-to-image models address pixel-aligned problems that largely preserve the input, such as inpainting~\cite{lugmayr2022repaint,saharia2022palette}, image restoration~\cite{xia2023diffir,conde2024instructir,saharia2022palette}, superresolution~\cite{li2022srdiff}, or depth estimation~\cite{saxena2023monocular}.
They shine on tasks where the output is pixel-aligned to the input and the task is clearly expressed in text form.
We instead focus on tasks where the output can and should vary drastically from the input, while the task cannot be clearly expressed in text form alone (\eg identity or style).

Palette~\cite{saharia2022palette} and InstructIR~\cite{conde2024instructir} show that a generalist model outperforms task-specific models.
While the former doesn't require instruction as its inputs are characteristic to each task, InstructIR~\cite{conde2024instructir} conditions on the noise model with text-based soft task routing --- similar to the cross attention to our proposed instruction, although our instructions are separated well enough in CLIP space (\cref{fig:tSNE-embedding-instructions}) that we do not require the dedicated projection layer like InstructIR.

For more high-level image translation tasks, InstructPix2Pix~\cite{brooks2023instructpix2pix} finetunes a base model to reinterpret its text prompt to perform \eg object replacement or style transfer, losing compatibility with pre-trained LoRAs and ControlNets in the process.
InstructPix2Pix is trained on Prompt2Prompt~\cite{hertz2023prompt2prompt} outputs, maximally preserving the input images in a pixel-aligned way.
InstructPix2Pix found multiple guidance necessary to handle the dual image and instruction conditions: IPAdapter and our proposed IPAdapterInstruct again benefit from a frozen base model, which allows a scale factor on the residual connection to control the influence of conditions.

\section{Method}\label{sec:method}%

\subsection{Preliminaries}\label{sec:preliminaries}%

Diffusion models~\cite{DiffusionTheory,DDPM,DDIM} iteratively generate images by gradually denoising pure noise --- they are trained to reverse a diffusion process that gradually transforms an image into noise (typically white Gaussian noise).
We write the forward noise process as starting from the data distribution $\vec{z}_0 \sim p(\vec{z})$ and ending with pure noise samples $\vec{z}_T \sim \mathcal{N}(0, \mathbf{I})$, over the course of $T$ time steps.
The immediate forward process is formally specified as
\begin{equation}
    \vec{z}_{t+1} \sim p(\vec{z}_{t+1} \vert \vec{z}_{t}) = \mathcal{N}\left(\sqrt{\alpha_{t+1}}\vec{z}_{t}, (1-\alpha_{t+1}) \mathbf{I}\right),
\end{equation}
where $\alpha_t$ denotes the so-called noise schedule.
Given this forward process, the diffusion model is trained to model the immediate denoising distributions, noted as $\hat{p}(\vec{z}_t \vert \vec{z}_{t+1})$.
During training, time steps are randomly sampled, and we directly supervise $\hat{p}(\vec{z}_t \vert \vec{z}_{t+1})$ by first sampling $p(\vec{z}_t \vert \vec{z}_0), \vec{z}_0 \sim p(\vec{z})$ followed by $\vec{z}_{t+1} \sim p(\vec{z}_{t+1}\vert \vec{z}_t)$.
Luckily, the exponential nature of additive white Gaussian noise implies a closed-form direct conditional $\vec{z}_t \sim p(\vec{z}_{t} \vert \vec{z}_0)$ which means that sampling $p(\vec{z}_t \vert \vec{z}_0)$ is constant-cost in terms of $t$.
By iteratively running the diffusion model for subsequent time steps, we can sample from the full generative model, written as
\begin{equation}
    \hat{p}(\vec{z}_0 \vert \vec{z}_{T}) = \prod\limits_{T}^{1}\hat{p}(\vec{z}_t-1 \vert \vec{z}_{t}), \qquad\vec{z}_{T} \sim \mathcal{N}(0, \mathbf{I}).
\end{equation}

As completely unconditional sampling is not very useful, diffusion models are trained to condition the generative distribution on an auxiliary input text prompt $\mathcal{T}$, modeling instead $p(\vec{z}_0 \vert \vec{z}_T, \mathcal{T})$, a technique known as \emph{Classifier-Free Guidance}~\cite{CFG}.
An IPAdapter model~\cite{IPAdapter} provides further controllability by additionally conditioning the generative model on a condition image $\mathcal{C}$ to model $p(\vec{z}_0 \vert \vec{z}_T, \mathcal{T}, \mathcal{C})$.
In practice, it leverages a pre-trained text-conditioned model and adds a cross-attention layer to the (projected) image condition after every prompt cross-attention layer.
The base model is kept frozen to preserve its generative performance and expressivity.
The condition image is encoded into the low-dimensional CLIP space~\cite{OpenAI-CLIP}.
This low-dimensional embedding does not contain pixel-accurate information, but rather a high-level semantic representation of the image with concepts of composition, style, subject, and object identities.

IPAdapters are trained by first sampling a data element $\vec{z}_0$ and then setting $\mathcal{C}=\vec{z}_0$.
\Ie{} the model is only supervised to (attempt to) reproduce the condition image exactly --- even though this is not the (only) intended use-case.
Although the model is never trained to produce images with a different caption than the condition image, it shows emergent capabilities to do exactly that: it tends to generate images with the same style, composition, and identities as the condition.
However, it lacks any controllability of these aspects, and sometimes fails any or all of these aspects, depending on the text prompt.

\subsection{Instruction-guided Image  Conditioning}%

As discussed in \cref{sec:related} and \cref{sec:preliminaries}, existing techniques that condition on images do not provide a clear way to control the condition's influence on the output: especially for natural condition images, there is no one way of incorporating their information and content.
Instead, we let the user clarify intent through an additional ``Instruct'' prompt $\mathcal{I}$, which we will refer to as the \emph{instruction} or \emph{instruct prompt} to distinguish it from the \emph{text prompt} $\mathcal{T}$.
This means we now  model the probability distribution $p(\vec{z}_0 \vert \vec{z}_T, \mathcal{T}, \mathcal{C}, \mathcal{I})$.
We consider the original IPAdapter~\cite{IPAdapter} an instantiation of this model where $\mathcal{I}=$\emph{``Reproduce everything from this image''} --- our approach models a wider posterior of which IPAdapter is a marginal.
In this paper we discuss five distinct generation tasks for the joint \emph{IPAdapter-Instruct} model:
\begin{itemize}
    \item\textbf{Replication}: variations of the condition (as IPAdapter),
    \item\textbf{Style}: an image in the style of the condition,
    \item\textbf{Composition}: an image with the same structure as the condition,
    \item\textbf{Object}: an image containing the object in the condition, and
    \item\textbf{Face}: an image containing the face of the person in the condition.
\end{itemize}

Given proper datasets and training  procedures, five separate IPAdapter instances could well handle each of these tasks: but that workflow is cumbersome, both at training time and at inference time.
Instead IPAdapter-Instruct is trained for all tasks simultaneously, making training the entire task set  more efficient and making inference more practical.
Furthermore, multi-task learning has proven to be beneficial in many contexts~\cite{Taskonomy}.
\cref{sec:architecture} outlines the model architecture in detail before \cref{sec:datasets} exhaustively outlines each of the above tasks and their training supervisiong.

\subsection{Model architecture}\label{sec:architecture}

Our model architecture is based on the transformer projection model from IPAdapter+~\cite{IPAdapter}.
We first discuss the original IPAdapter+ architecture, followed by our modifications.

\subsubsection{IPAdapter+} adds a cross-attention layer to a projected encoding of the condition image after every cross-attention to the text prompt $\mathcal{T}$, as shown in \cref{fig:ipadapter-injection}.
The condition image is first encoded to the CLIP domain and then projected to an IPAdapter+-specific space with single linear layer before passing through a small transformer model as shown in \cref{fig:ipadapterplus}.

\subsubsection{Introducing the instruction.}
For IPAdapter-Instruct, we've modified the projection transformer model to introduce an additional attention layer at every iteration that also attends to the CLIP embedding of the instruction, as shown in \cref{fig:ipadapterinstruct}.
In this way, the model is granted the capacity to extract the relevant information (per the instruction) from the condition embedding. 

\begin{figure}[ht]
    \begin{center}
    \def\aimbgsize{0.35}
\resizebox{\linewidth}{!}{
\begin{tikzpicture}[
    node distance=1cm, 
    font={\fontsize{12pt}{10}\selectfont},
    layer/.style={draw=#1, fill=#1!20, line width=1.5pt, inner sep=0.2cm, rounded corners=1pt},
    textlayer/.style={draw=#1, fill=#1!20, line width=1.5pt, inner sep=0.1cm, rounded corners=1pt},
    encoder/.style={isosceles triangle,isosceles triangle apex angle=60,shape border rotate=0,anchor=apex,draw=#1, fill=#1!20, line width=1.5pt, inner sep=0.1cm, rounded corners=1pt},
    decoder/.style={isosceles triangle,isosceles triangle apex angle=60,shape border rotate=180,anchor=apex,draw=#1, fill=#1!20, line width=1.5pt, inner sep=0.1cm, rounded corners=1pt},
    label/.style={font=\scriptsize, text width=1.2cm, align=center},
    mynode/.style={align=center,draw=blueish, fill=white, rounded corners=1pt, line width=1.5pt},
    datanode/.style={align=center,draw=blueish, fill=white, rounded corners=1pt, line width=1.5pt},
    learnnode/.style={align=center,draw=blueish, fill=blueish!20, rounded corners=1pt, line width=1.5pt},
    instructnode/.style={align=center,draw=orangeish, fill=orangeish!20, rounded corners=1pt, line width=1.5pt}
]

\begin{scope}[shift={(0,0)},rotate=90,scale=3,local bounding box=unet-scope]
    \node[draw=blueish, fill=white, text width=1.25cm, text height=0.1cm, rounded corners=1pt, line width=1.5pt,transform shape] (rgb-unet-1) at (0, 0) {};
    \node[draw=blueish, fill=white, text width=0.7cm, text height=0.1cm, below=0.1cm of rgb-unet-1, rounded corners=1pt, line width=1.5pt,transform shape] (rgb-unet-2) {};
    \node[draw=blueish, fill=white, text width=0.25cm, text height=0.1cm, below=0.1cm of rgb-unet-2, rounded corners=1pt, line width=1.5pt,transform shape] (rgb-unet-3) {};
    \node[draw=blueish, fill=white, text width=0.25cm, text height=0.1cm, below=0.1cm of rgb-unet-3, rounded corners=1pt, line width=1.5pt,transform shape] (rgb-unet-4) {};
    \node[draw=blueish, fill=white, text width=0.7cm, text height=0.1cm, below=0.1cm of rgb-unet-4, rounded corners=1pt, line width=1.5pt,transform shape] (rgb-unet-5) {};
    \node[draw=blueish, fill=white, text width=1.25cm, text height=0.1cm, below=0.1cm of rgb-unet-5, rounded corners=1pt, line width=1.5pt,transform shape] (rgb-unet-6) {};
    
    \begin{scope}[on background layer]
        \fill[fill=blueish!20, rounded corners=0pt] 
            ($(rgb-unet-1.north west)+(-0.275,0.15)$) -- 
            ($(rgb-unet-3.west)!0.5!(rgb-unet-4.west)+(-0.275,0)$) -- 
            ($(rgb-unet-6.south west)+(-0.275,-0.15)$) -- 
            ($(rgb-unet-6.south east)+(0.275,-0.15)$) -- 
            ($(rgb-unet-3.east)!0.5!(rgb-unet-4.east)+(0.275,0)$) -- 
            ($(rgb-unet-1.north east)+(0.275,0.15)$) -- cycle;%
    \end{scope}
\end{scope}

\node[learnnode,align=center,anchor=south] (text-attention) at ($(rgb-unet-2.east) + (0.0,3)$) {Text prompt\\Cross-Attention};
\node[instructnode,align=center,anchor=south] (ip-attention) at ($(rgb-unet-5.east) + (0.0,3)$)  {IPAdapter\\Cross-Attention};

\node[font=\fontsize{20}{20},text=blueish] at ($(text-attention.north east) + (-0.1,-0.05)$) {\faSnowflake};
\node[font=\fontsize{20}{20},text=red] at ($(ip-attention.north east) + (-0.1,-0.05)$) {\faGripfire};
\node[font=\fontsize{20}{20},text=blueish] at ($(unet-scope.south east)$) {\faSnowflake};

\def\aimarrowoffset{0.2}
\foreach \i in {1,2,...,6} {
\draw[-stealth,line width=1.5pt,draw=blueish] ($(text-attention.south) + (-\aimarrowoffset,0.0)$) --  node {} ++(0,-0.8-\aimarrowoffset) -| ($(rgb-unet-\i.east) + (-\aimarrowoffset,0.0)$);
}
\foreach \i in {1,2,...,6} {
\draw[-stealth,line width=1.5pt,draw=orangeish] ($(ip-attention.south) + (\aimarrowoffset,0.0)$) --  node {} ++(0,-0.8+\aimarrowoffset) -| ($(rgb-unet-\i.east) + (\aimarrowoffset,0.0)$);
}

\node[datanode,anchor=east,align=center] (input-text) at ($(text-attention.west) + (-2.5,0.0)$) {Text Prompt\\Features};
\node[datanode,anchor=west,align=center] (input-condition) at ($(ip-attention.east) + (2.5,0.0)$) {Text Condition\\Features};

\draw[-stealth,line width=1pt] (input-text) -- (text-attention);
\draw[-stealth,line width=1pt] (input-condition) -- (ip-attention);

\node[left=of unet-scope,font=\fontsize{16}{16}] (noisy-latents) {$\vec{z}-{t+1}$};
\node[right=of unet-scope,font=\fontsize{16}{16}] (denoised-latents) {$\vec{z}-{t}$};

\draw[-stealth,line width=1pt] (noisy-latents) -- (rgb-unet-1);
\draw[-stealth,line width=1pt] (rgb-unet-6) -- (denoised-latents);

\end{tikzpicture}
}
    \end{center}
    \caption{IPAdapter(+) and IPAdapter-Instruct both inject the conditioning in the same way, using an additional cross-attention layer after every text prompt cross attention layer. Only the way the condition is condensed into conditioning features differs, as discussed in \cref{sec:architecture} and shown in \cref{fig:overview}.
    The original network weights (including the text prompt cross-attention) are kept frozen and only the new cross-attention is being learned.
    }\label{fig:ipadapter-injection}
\end{figure}
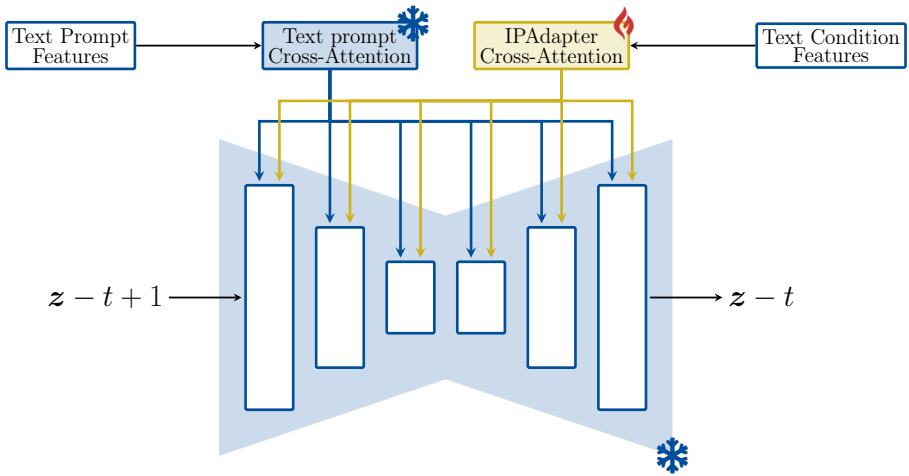

\begin{figure}[ht]
    \centering
    \input{fig/architecture_single.tex}
\end{figure}

\subsubsection{Encoding the instruction.}
We have chosen to encode the instruction using a text embedding model, embedding it into the same space the condition image is being embedded into by the original IPAdapter(+) model.
Given the discrete nature of our task set, we could also learn task-specific embeddings for each of the tasks, and use those instead.
However, by leveraging the powerful pre-trained ViT-H/14~\cite{ViT-H/14CLIP} model, we benefit from the semantic richness of the CLIP embedding space and from representing both the instruction and the condition in the same space.
Although we do not investigate it in detail in this work, our intuition is that this leads to more flexible and robust instruction understanding and provides a better starting point for additional future tasks.

\subsection{Tasks and Dataset Generation}\label{sec:datasets}

\begin{figure}[t]
    \begin{subfigure}[b]{0.19\linewidth}
        \centering
        \includegraphics[width=\linewidth,height=\linewidth]{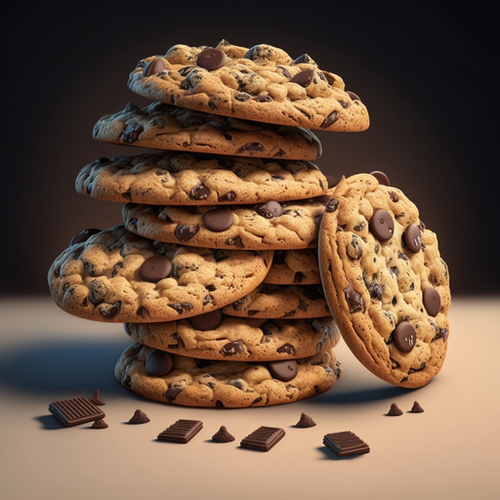}
        \begin{tikzpicture}
            \draw[-stealth,line width=1pt]                  (0,1.5)   -- (0,0.75) node [fill=white,align=center,font=\tiny] {\parbox{0.9\linewidth}{\centering{}\input{img/paper_imagery/fixed/dataset_everything_instruct.txt}}}  -- (0,0) node [anchor=north,font=\tiny,align=center]{\parbox{0.9\linewidth}{\centering{}\input{img/paper_imagery/fixed/dataset_everything_prompt.txt}}};
        \end{tikzpicture}
        \includegraphics[width=\linewidth,height=\linewidth]{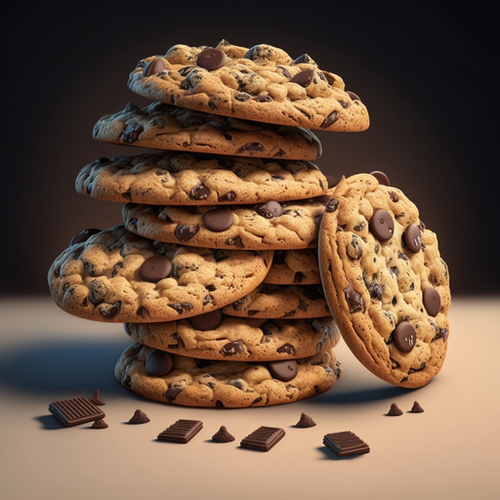}
        \caption[Image replication]{\centering\parbox{\textwidth}{\centering Image replication}}\label{fig:datasets:everything}
    \end{subfigure}
    \hfill
    \begin{subfigure}[b]{0.19\linewidth}
        \centering
        \includegraphics[width=\linewidth,height=\linewidth]{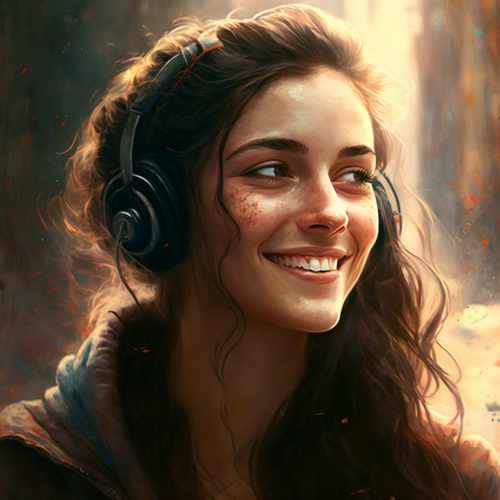}
        \begin{tikzpicture}
            \draw[-stealth,line width=1pt]                  (0,1.5)   -- (0,0.75) node [fill=white,align=center,font=\tiny] {\parbox{0.9\linewidth}{\centering{}\input{img/paper_imagery/fixed/dataset_style_instruct.txt}}}  -- (0,0) node [anchor=north,font=\tiny,align=center]{\parbox{0.9\linewidth}{\centering{}\input{img/paper_imagery/fixed/dataset_style_prompt.txt}}};
        \end{tikzpicture}
        \includegraphics[width=\linewidth,height=\linewidth]{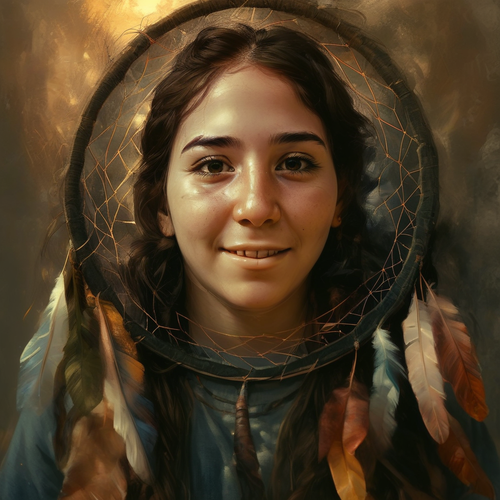}
        \caption[Style transfer]{\centering\parbox{\textwidth}{\centering Style transfer}}\label{fig:datasets:style}
    \end{subfigure}
    \hfill
    \begin{subfigure}[b]{0.19\linewidth}
        \centering
        \includegraphics[width=\linewidth,height=\linewidth]{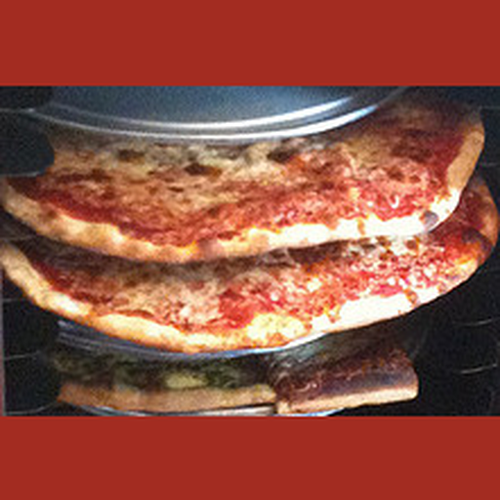}
        \begin{tikzpicture}
            \draw[-stealth,line width=1pt]                  (0,1.5)   -- (0,0.75) node [fill=white,align=center,font=\tiny] {\parbox{0.9\linewidth}{\centering{}\input{img/paper_imagery/fixed/dataset_object_instruct.txt}}}  -- (0,0) node [anchor=north,font=\tiny,align=center]{\parbox{0.9\linewidth}{\centering{}\input{img/paper_imagery/fixed/dataset_object_prompt.txt}}};
        \end{tikzpicture}
        \includegraphics[width=\linewidth,height=\linewidth]{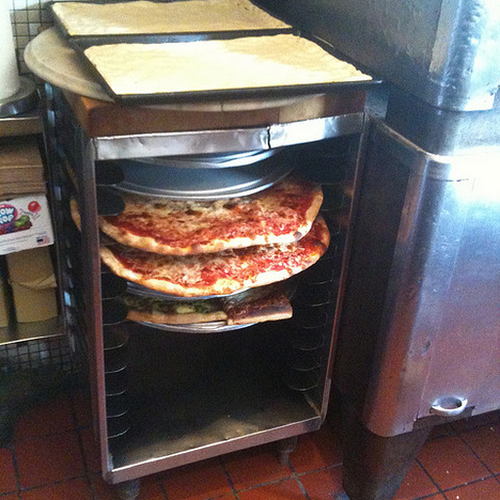}
        \caption[Object extraction]{\centering\parbox{\textwidth}{\centering Object extraction}}\label{fig:datasets:object}
    \end{subfigure}
    \hfill
    \begin{subfigure}[b]{0.19\linewidth}
        \centering
        \includegraphics[width=\linewidth,height=\linewidth]{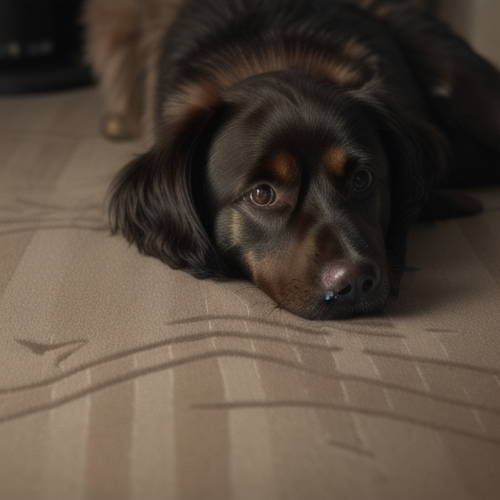}
        \begin{tikzpicture}
            \draw[-stealth,line width=1pt]                  (0,1.5)   -- (0,0.75) node [fill=white,align=center,font=\tiny] {\parbox{0.9\linewidth}{\centering{}\input{img/paper_imagery/fixed/dataset_composition_instruct.txt}}}  -- (0,0) node [anchor=north,font=\tiny,align=center]{\parbox{0.9\linewidth}{\centering{}\input{img/paper_imagery/fixed/dataset_composition_prompt.txt}}};
        \end{tikzpicture}
        \includegraphics[width=\linewidth,height=\linewidth]{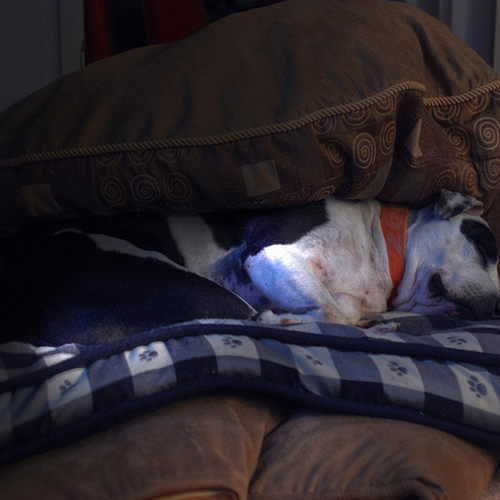}
        \caption[Composition]{\centering\parbox{\textwidth}{\centering Composition}}\label{fig:datasets:composition}
    \end{subfigure}
    \hfill
    \begin{subfigure}[b]{0.19\linewidth}
        \centering
        \includegraphics[width=\linewidth,height=\linewidth]{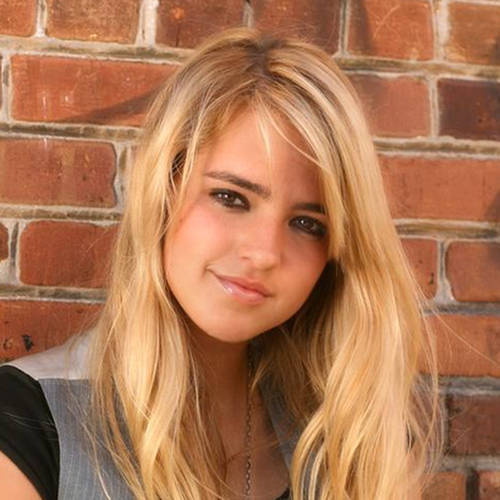}
        \begin{tikzpicture}
            \draw[-stealth,line width=1pt]                  (0,1.5)   -- (0,0.75) node [fill=white,align=center,font=\tiny] {\parbox{0.9\linewidth}{\centering{}\input{img/paper_imagery/fixed/dataset_face_instruct.txt}}}  -- (0,0) node [anchor=north,font=\tiny,align=center]{\parbox{0.9\linewidth}{\centering{}\input{img/paper_imagery/fixed/dataset_face_prompt.txt}}};
        \end{tikzpicture}
        \includegraphics[width=\linewidth,height=\linewidth]{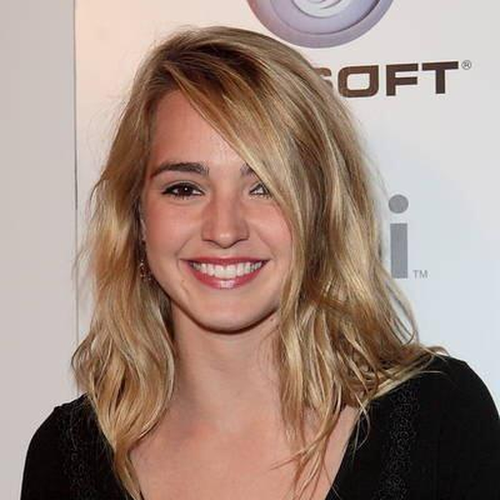}
        \caption[Face extraction]{\centering\parbox{\textwidth}{\centering Face extraction}}\label{fig:datasets:face}
    \end{subfigure}
    
    \caption{Example dataset entries for each of the five tasks. Instructions and prompts are shortened for print.}\label{fig:datasets}
\end{figure}

We build a dedicated dataset for each of the distinct tasks, which  we discuss in detail below.
For the instruction prompts, we generate example instruct prompts for each task using a large language model (LLM)~\cite{ChatGPT4}: these prompts are randomly sampled during the training procedure.
To ensure that each task is well identifiable in CLIP space, we assign a keyword to each task and remove any instructions that contain a keyword from another task.
\Cref{fig:tSNE-embedding-instructions}  shows a t-SNE visualization of the instruction's embeddings for each of the tasks (as well as the ``average'' instruction used in the ablation studies).
Example entries from all of these datasets are shown in \cref{fig:datasets}.

\subsubsection{Image replication}
As in IPAdapter's training procedure, the goal is to create slight but subtle variations of the input image.
As this mode replicates IPAdapter's original behavior, we expect the same emergent capability to re-use this mode of instruction for other tasks.
To create the training dataset, we use JourneyDB~\cite{JourneyDB} dataset, collecting $42,000$ random examples with their original text prompts.
The instruction prompts are generated by querying ChatGPT4~\cite{ChatGPT4} to ``\emph{Generate ways to describe taking everything from an image of varying lengths, do not use the words composition, style, face or object}''.

\subsubsection{Style Preservation}
In style preservation, the user wants to extract only the style of the condition image and apply it to a new image --- although this is not well-defined, it is taken to encompass the color scheme and general art style.
Identities, layout and composition are not intended to leak into the generation.
In order to create the style training dataset, we start from the \emph{ehristoforu/midjourney-images}~\cite{MidjourneyImages} style dataset and a large-scale art dataset~\cite{ArtDataset}.
The condition images are sampled from the style dataset, whereas the target images are created using InstantStyle~\cite{InstantStyle} with prompts from the art dataset and the condition as the style source.
\Ie this mode is supervised to generate images as InstantStyle would.
The instruction prompts are generated by querying ChatGPT4 to ``\emph{Generate ways to describe taking the style from an image of varying lengths, do not use the words composition, object, face or everything}''.
This dataset consists only of $20000$ examples, as InstantStyle is computationally expensive and slow.

\subsubsection{Object Extraction}
Here, the goal is to place the object from the condition in a new scene, preserving its identity as much as possible, similar to MagicInsert~\cite{MagicInsert}.
We generate this dataset based on COCO~\cite{COCO}: for $35000$, as condition, we crop out the relevant object and (if necessary) pad with a random color.
The target image is the original dataset image, for which a text prompt is provided by GPT4o~\cite{GPT4o}.
The instruction prompts are generated by querying ChatGPT4 to ``\emph{Generate ways to describe taking an object from an image of varying lengths, do not use the words composition, style, face or everything}''.
To provide more semantic information (which the typical user will have access to) to the image projection layer from \cref{fig:ipadapterinstruct}, we replace the `object' keyword in the instruction with the name of the object.

\subsubsection{Structural Preservation}
Finally, we also create a dataset for structural preservation.
This is intended to replicate the behavior of a scribble ControlNet~\cite{ControlNet} model, which generates images with similar canny edge profiles as the condition, but without having to explicitly generate those edge images first.
For the scribble dataset, we use the CommonCanvas dataset~\cite{CommonCanvas} and generate new target images using the \emph{lllyasviel/sd-controlnet-scribble} ControlNet from their canny edge maps and original prompts--- the original images themselves are used as the condition.
The instruction prompts are generated by querying ChatGPT4 to ``\emph{Generate ways to describe taking the composition from an image of varying lengths, do not use the words style, object, face or everything}''.

\subsubsection{Identity Preservation}
As humans are extremely sensitive to facial features, we also create a dedicated dataset for face preservation using CelebA~\cite{CelebA}.
We sampled $40000$ matching image pairs, and GPT4o~\cite{GPT4o} provided the text prompts.
Half of the condition images were zoomed in on the face in order to focus even more on facial features.
The instruction prompts are generated by querying ChatGPT4 to ``\emph{Generate ways to describe taking the face or identity from an image of varying lengths, do not use the words composition, style, everything or object}''.

\subsection{Training Process}

Our training process follows IPAdapter~\cite{IPAdapter} training procedure, using the datasets discussed in \cref{sec:datasets}.
The base model of choice is StableDiffusion 1.5~\cite{LatentDiffusionModels} for its excellent balance of output diversity, controllability and accessibility --- it remains a staple model in the community for these reasons.
The base model remains fully frozen, and we initialize the original IPAdapter elements from the IPAdapter+ weights for SD1.5~\cite{IPAdapterPlusSD15}.
Most of the new residual layers are initialized with white noise ($\sigma=10^{-4}$).
The final activation are zero-initialized to initialize replicating the base IPAdapter.

For the main model, we use a batch size of $512$ and a learning rate of $10^{-6}$ for a  total of $100000$ steps.
For the ablation studies, to stymie the cost, we use a batch size of $64$ and a learning rate of $10^{-7}$ for a total of $100000$ steps --- visually, the results have already converged and further training is unlikely to impact our ablation conclusions.
Similar to IPAdapter~\cite{IPAdapter}, we find that for user inference the impact of the residual connection from the IPAdapter needs to be scaled back 20\%-40\%, to avoid overpowering the text cross attention.

\section{Experiments and results}\label{sec:results}

\subsection{Datasets and metrics}
\Cref{sec:method} already discussed the dataset creation and training procedures in detail.
For each of the task datasets, we hold out a validation set of $1000$ images  for quantitative and qualitative evaluations.
We use the following metrics for evaluating each of the tasks:
\begin{itemize}
\item \textbf{CLIP-I}~\cite{IPAdapter}: cosine similarity between the generated image and the condition, to indicate how much information was passed from the condition to the generation.
\item \textbf{CLIP-T}~\cite{IPAdapter}: ClipScore~\cite{CLIPScore} between the generated image and the original caption of the condition, to indicate the success of the replication task.
\item \textbf{CLIP-P}: ClipScore~\cite{CLIPScore} between the generated image and the user text prompt, to indicate how well the text prompt was followed for the non-replication tasks.
\item \textbf{CLIP Style Score (CLIP-S)}: cosine similarity between (a) the generated image's CLIP embedding minus its text prompt's CLIP embedding and (b) the condition image's CLIP embedding minus its known text prompt's embedding, motivated by InstantStyle~\cite{InstantStyle}'s performance, to indicate style transfer success.
\end{itemize}

\subsection{Compared to task-specific models}

\begin{figure}[t]
    \centering
    \newlength{\aimqualsubfigsize}
\setlength{\aimqualsubfigsize}{0.49\linewidth}
\newlength{\aimqualimgsize}
\setlength{\aimqualimgsize}{0.28\aimqualsubfigsize}
\newlength{\aimqualtxtsize}
\setlength{\aimqualtxtsize}{0.055\aimqualsubfigsize}
\centering

\resizebox{\aimqualsubfigsize}{!}{
\tiny
\begin{tabular}{>{\centering\arraybackslash}m{\aimqualimgsize}>{\centering\arraybackslash}m{\aimqualtxtsize}>{\centering\arraybackslash}m{\aimqualimgsize}>{\centering\arraybackslash}m{\aimqualimgsize}}
    Style & & IPAdapter-Instruct & Single-task \\\centering%
    \includegraphics[width=0.8\linewidth,height=0.8\linewidth]{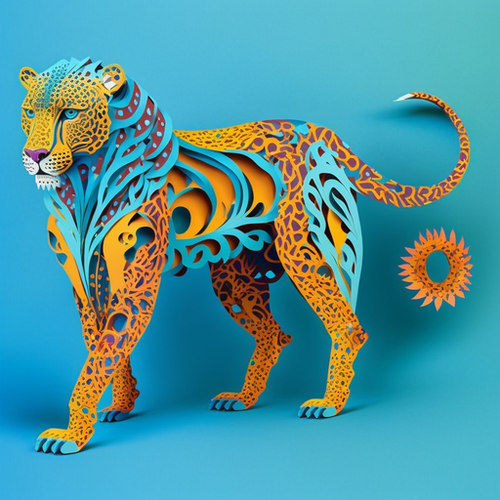}&\centering%
    \rotatebox{90}{\parbox{0.9\aimqualimgsize}{Profile of a cat}}&%
    \includegraphics[width=\linewidth,height=\linewidth]{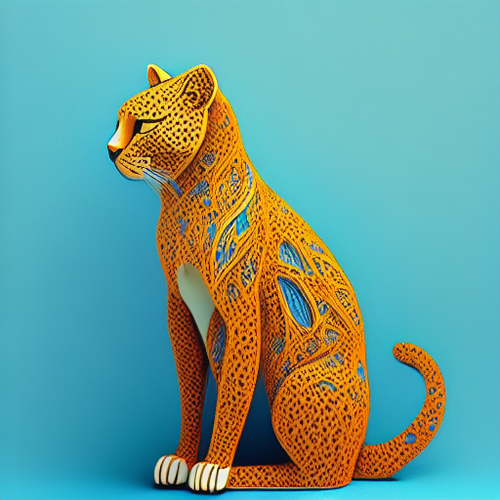}&%
    \includegraphics[width=\linewidth,height=\linewidth]{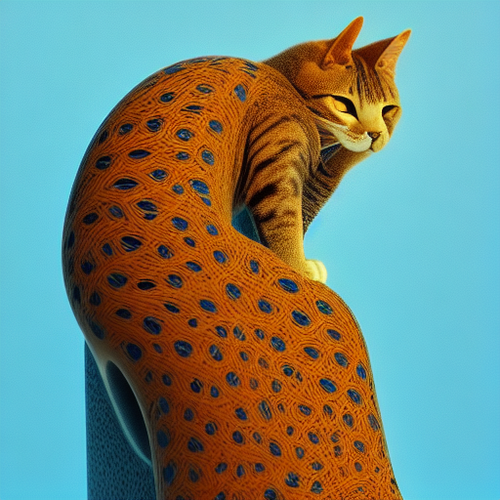}\\\centering%
    \includegraphics[width=0.8\linewidth,height=0.8\linewidth]{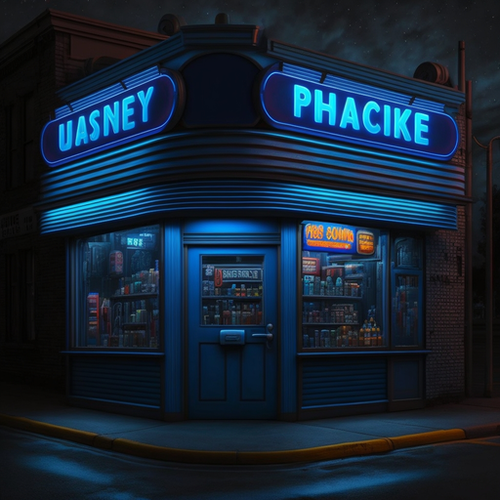}&\centering%
    \rotatebox{90}{\parbox{0.9\aimqualimgsize}{Red-green\\storefront}}&%
    \includegraphics[width=\linewidth,height=\linewidth]{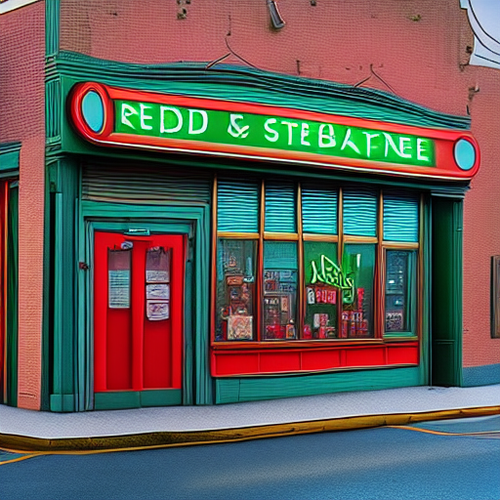}&%
    \includegraphics[width=\linewidth,height=\linewidth]{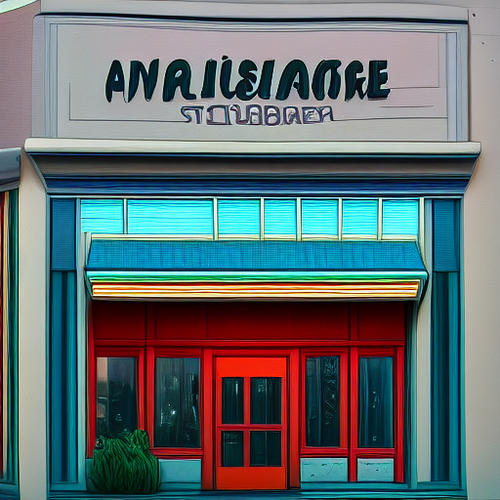}
\end{tabular}
}
\hfill
\resizebox{\aimqualsubfigsize}{!}{
\tiny
\begin{tabular}{>{\centering\arraybackslash}m{\aimqualimgsize}>{\centering\arraybackslash}m{\aimqualtxtsize}>{\centering\arraybackslash}m{\aimqualimgsize}>{\centering\arraybackslash}m{\aimqualimgsize}}
    Object & & IPAdapter-Instruct & Single-task\\\centering%
    \includegraphics[width=0.8\linewidth,height=0.8\linewidth]{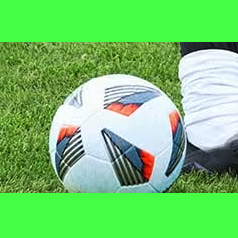}&\centering%
    \rotatebox{90}{\parbox{0.9\aimqualimgsize}{Playing soccer}}&%
    \includegraphics[width=\linewidth,height=\linewidth]{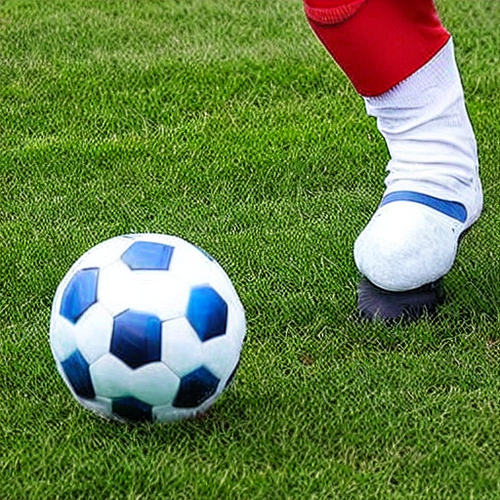}&%
    \includegraphics[width=\linewidth,height=\linewidth]{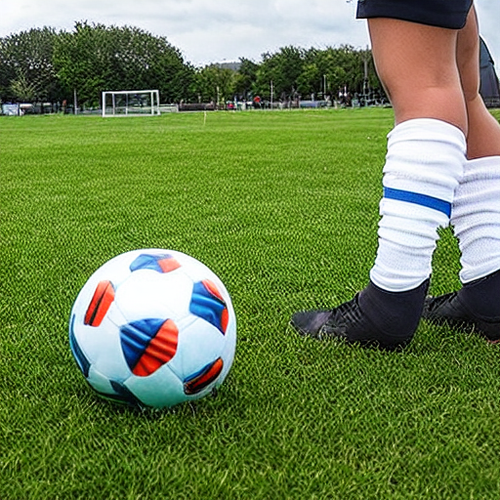}\\\centering%
    \includegraphics[width=0.8\linewidth,height=0.8\linewidth]{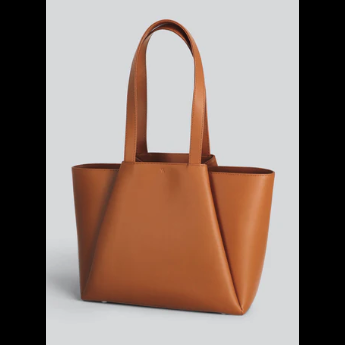}&\centering%
    \rotatebox{90}{\parbox{0.9\aimqualimgsize}{A woman's\\handbag}}&%
    \includegraphics[width=\linewidth,height=\linewidth]{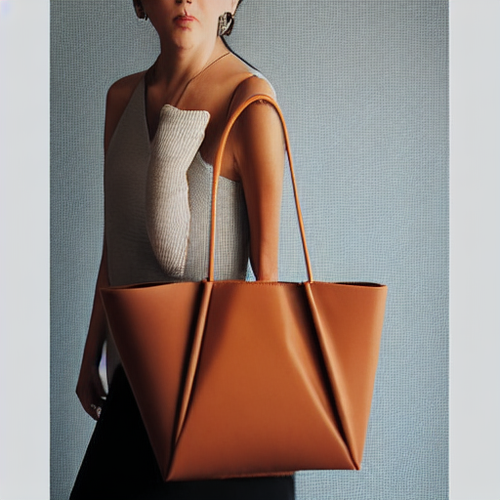}&%
    \includegraphics[width=\linewidth,height=\linewidth]{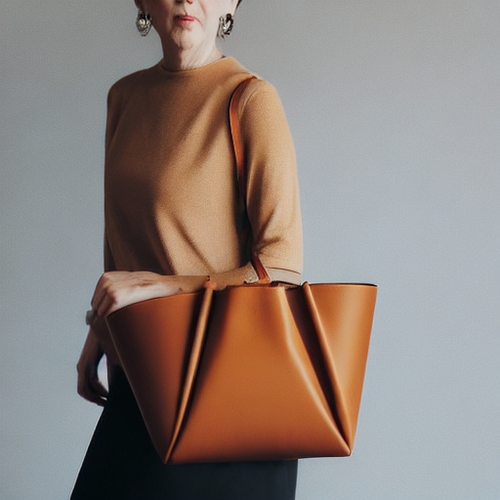}
\end{tabular}
}
\vspace*{0.5\baselineskip}

\resizebox{\aimqualsubfigsize}{!}{
\tiny
\begin{tabular}{>{\centering\arraybackslash}m{\aimqualimgsize}>{\centering\arraybackslash}m{\aimqualtxtsize}>{\centering\arraybackslash}m{\aimqualimgsize}>{\centering\arraybackslash}m{\aimqualimgsize}}
    Composition & & IPAdapter-Instruct & Single-task\\\centering%
    \includegraphics[width=0.8\linewidth,height=0.8\linewidth]{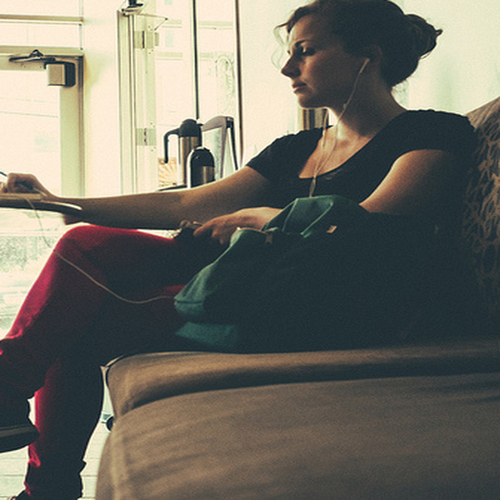}&\centering%
    \rotatebox{90}{\parbox{0.9\aimqualimgsize}{A girl and dog\\in the couch}}&%
    \includegraphics[width=\linewidth,height=\linewidth]{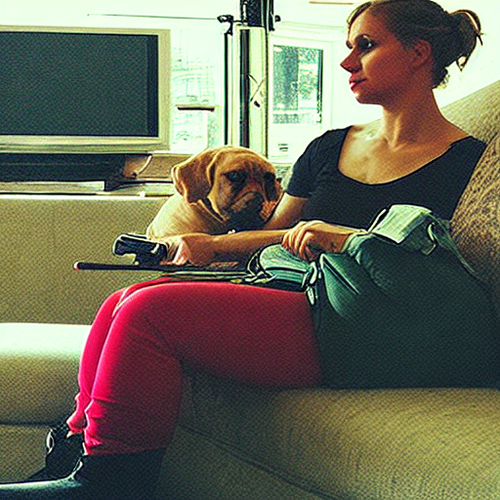}&%
    \includegraphics[width=\linewidth,height=\linewidth]{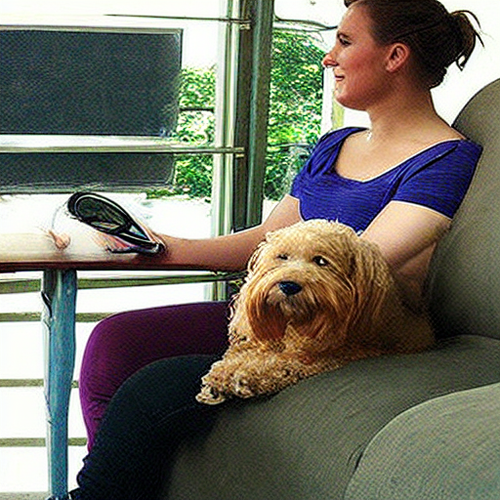}\\\centering%
    \includegraphics[width=0.8\linewidth,height=0.8\linewidth]{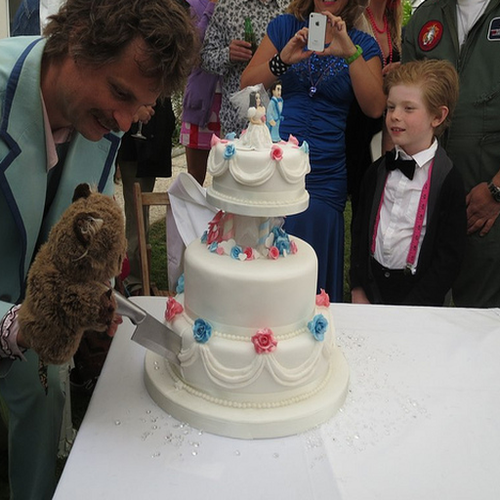}&\centering%
    \rotatebox{90}{\parbox{0.9\aimqualimgsize}{A pink\\unicorn cake}}&%
    \includegraphics[width=\linewidth,height=\linewidth]{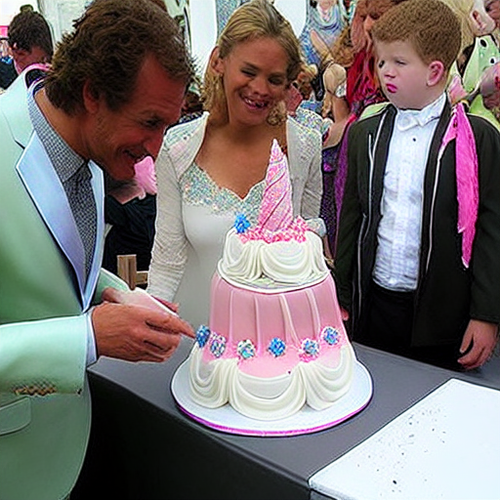}&%
    \includegraphics[width=\linewidth,height=\linewidth]{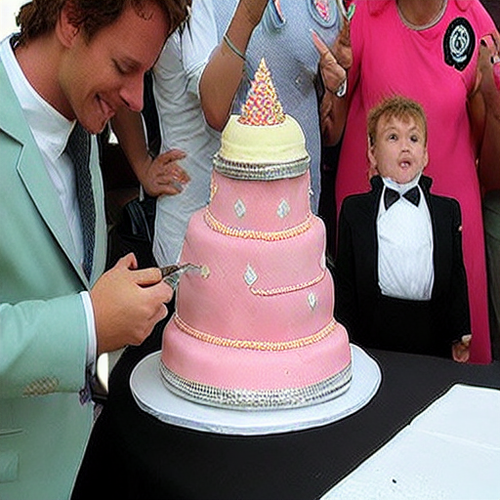}
\end{tabular}
}
\hfill
\resizebox{\aimqualsubfigsize}{!}{
\tiny
\begin{tabular}{>{\centering\arraybackslash}m{\aimqualimgsize}>{\centering\arraybackslash}m{\aimqualtxtsize}>{\centering\arraybackslash}m{\aimqualimgsize}>{\centering\arraybackslash}m{\aimqualimgsize}}
    Face & & IPAdapter-Instruct & Single-task\\\centering%
    \includegraphics[width=0.8\linewidth,height=0.8\linewidth]{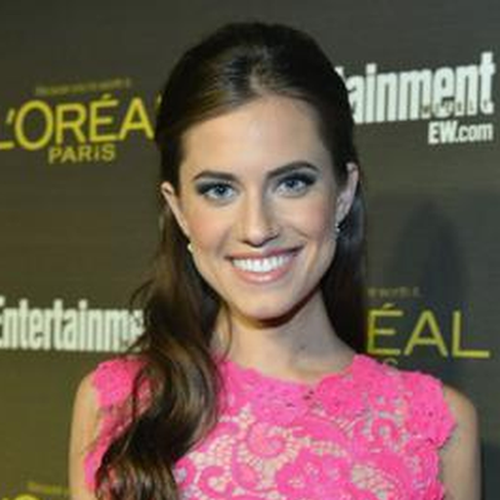}&\centering%
    \rotatebox{90}{\parbox{0.9\aimqualimgsize}{a woman\\carrying a dog}}&%
    \includegraphics[width=\linewidth,height=\linewidth]{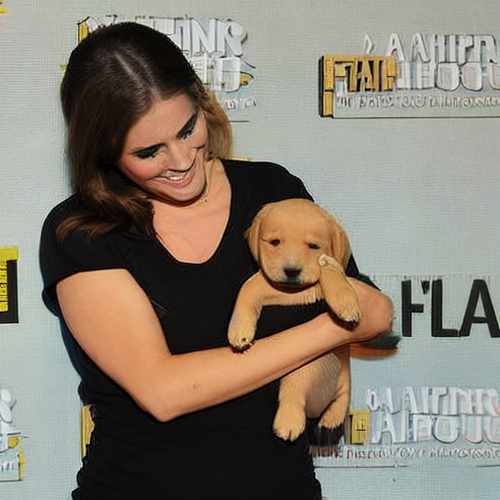}&%
    \includegraphics[width=\linewidth,height=\linewidth]{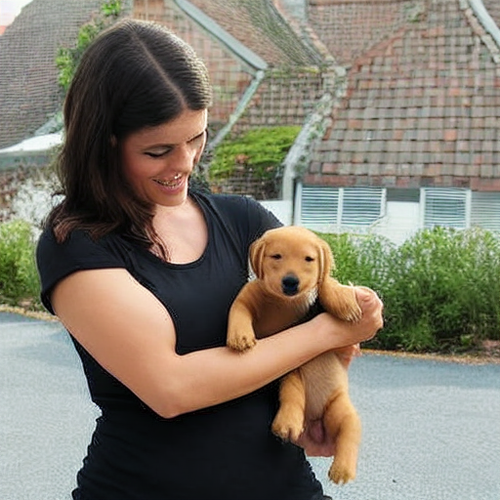}\\\centering%
    \includegraphics[width=0.8\linewidth,height=0.8\linewidth]{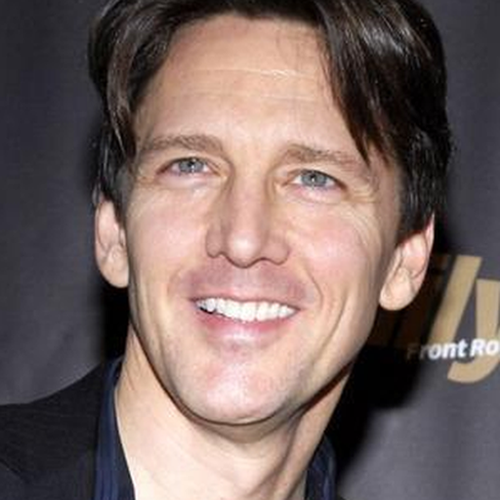}&\centering%
    \rotatebox{90}{\parbox{0.9\aimqualimgsize}{A clown with\\painted face}}&%
    \includegraphics[width=\linewidth,height=\linewidth]{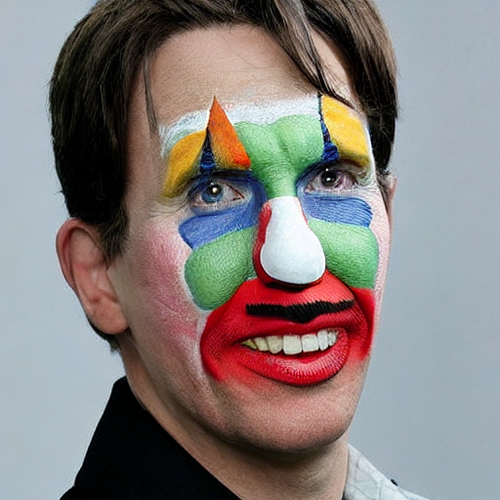}&%
    \includegraphics[width=\linewidth,height=\linewidth]{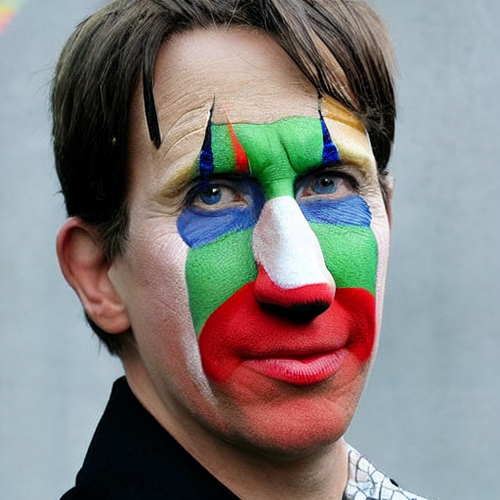}
\end{tabular}
}

    \caption{Qualitative comparison of our proposed IPAdapter-Instruct model to task-specific models. We omit the \emph{Replication} task as we compare against the base IPAdapter+ model in \cref{fig:vs-ipadapter-single-image}. Performance is similar, but occasionally the joint model leaks unintended aspects --- \cref{sec:qual-results} shows that further training resolves this.}\label{fig:vs-singletask-qualitative}
\end{figure}

We compare the performance of our model to that of models that were trained on each of the tasks specifically.
Please refer \cref{fig:vs-singletask-qualitative} for qualitative examples of the different models, and to \cref{tab:vs-singletask-quantitative} for a quantitative overview.

We see that our proposed model is on-par with or slightly better than the single-task models, while condensing them into a single model.
As \cref{fig:vs-singletask-speed} shows, the model trains just as fast as the single-task models, but for all tasks simultaneously. This drastically decreases total training time and cost for the entire task set --- aside from simplifying inference code handling.

\begin{figure}[t]
    \centering
    \begin{minipage}[c]{\linewidth}
    \centering
    \captionof{table}{Quantitative comparison of our IPAdapter-Instruct model to task-specific models, and to a model trained and evaluated with hardcoded queries. Very little quantitative difference is found so that we prefer the single joint model with flexible instruction queries. Higher is better for all of these metrics.}\label{tab:vs-singletask-quantitative}
    \resizebox{\linewidth}{!}{
    \begin{tabular}{rc:c|c:c|c:c|c:c|c:c}
       & \multicolumn{2}{c|}{Replication}                         & \multicolumn{2}{c|}{Style}                               & \multicolumn{2}{c|}{Object}                              & \multicolumn{2}{c|}{Composition}                         & \multicolumn{2}{c}{Faces}                               \\[5pt]
                   & \multicolumn{1}{c}{CLIP-I} & \multicolumn{1}{c|}{CLIP-T} & \multicolumn{1}{c}{CLIP-S} & \multicolumn{1}{c|}{CLIP-P} & \multicolumn{1}{c}{CLIP-I} & \multicolumn{1}{c|}{CLIP-P} & \multicolumn{1}{c}{CLIP-I} & \multicolumn{1}{c|}{CLIP-P} & \multicolumn{1}{c}{CLIP-I} & \multicolumn{1}{c}{CLIP-P} \\ \cline{2-11}
Single-task Model\hspace{3mm}      & 0.9152	&	0.8523	&	1.0988	&	0.7615	&	0.7670	&	0.6351	&	0.7337	&	0.6424	&	0.5154	&	0.6012\\
IPAdapter-Instruct\hspace{3mm}     & 0.9098	&	0.8542	&	1.1171	&	0.7776	&	0.8040	&	0.6402	&	0.7082	&	0.6517	&	0.5264	&	0.5976\\
Fixed instruct prompts\hspace{3mm} & 0.9086	&	0.8528	&	1.1141	&	0.7810	&	0.8029	&	0.6408	&	0.7151	&	0.6517	&	0.5159	&	0.6009
\end{tabular}

}
\vspace{2mm}
    \end{minipage}

    \begin{minipage}[c]{0.49\linewidth}
    \centering
    \input{fig/vs_singletask_training_speed.tex}
    \end{minipage}\setcounter{subfigure}{0}%
    \hfill
    \begin{minipage}[c]{0.49\linewidth}
    \begin{center}
    \input{fig/prompt_variation.tex}
    \end{center}
    \end{minipage}\setcounter{subfigure}{0}%

    \begin{minipage}[t]{0.49\linewidth}
    \captionof{figure}{We show the evolution of validation scores in function of training step for the (a) style and (b) face tasks, and note that the joint model's training is on par with the single-task training speed.}\label{fig:vs-singletask-speed}
    \end{minipage}
    \hfill
    \begin{minipage}[t]{0.49\linewidth}
    \captionof{figure}{Rewording the instruction prompt for style copy (a, b) and face transfer (c, d), drawn randomly from validation instruct prompts, shows a much more subtle impact than varying initial noise.}\label{fig:prompt variation inference}
    \end{minipage}
\end{figure}

\subsection{Compared to a fixed instruction set}

We ablate the choice to generate random prompts for the individual tasks rather than use a single, hard-coded, instruction prompt for each task.
In the latter case, \cref{tab:vs-singletask-quantitative} shows that the random prompts are more effective than the fixed prompts --- we intuit that this is because the model is forced to leverage similarities between the various tasks to some extent, when various wordings of different tasks are close-by in the prompt embedding space.
Furthermore, we show in \cref{fig:prompt variation inference} that the effect of varying the instruction for a given task has only a minimal, if noticeable, effect, allowing for minor output exploration.

\subsection{Qualitative results for the main model}\label{sec:qual-results}

Finally, \cref{fig:vs-ipadapter-single-image} provides qualitative results for our proposed approach against IPAdapter+ and InstructPix2Pix.
We see that it is hard to guide IPAdapter+ towards specific use-cases, whereas our model is easily controllable through the instruction.
It succeeds only when the user intent is clear: when the input image and prompt match perfectly the user requires a slight variation of the condition, and when they do not match at all the user requires style transfer.
It is also suboptimal at face identity, although our method shows that the CLIP space can be pushed much further.
InstructPix2Pix\footnote{We evaluate only the available pretrained network~\cite{brooks2023instructpix2pix}. Retraining InstructPix2Pix with our datasets is an option, but we consider it out of scope for this manuscript.} struggles when the output should differ significantly from the input (as expected) and neither can it retain facial identity, but it shines when only minor additions or edits to the input are required.
We find that its general output quality is slightly lower, which we attribute to fine-tuning versus freezing the base model.

\noindent\begin{minipage}{\linewidth}
\begin{center}
\newlength{\aimvsipadaptertasksimgsize}
\setlength{\aimvsipadaptertasksimgsize}{4.2cm}
\resizebox{0.9\linewidth}{!}{
    \renewcommand{\arraystretch}{0.6}%
    \setlength{\tabcolsep}{2pt}%
    \begin{tabular}{cccccc}%
        &%
        \includegraphics[width=0.7\aimvsipadaptertasksimgsize]{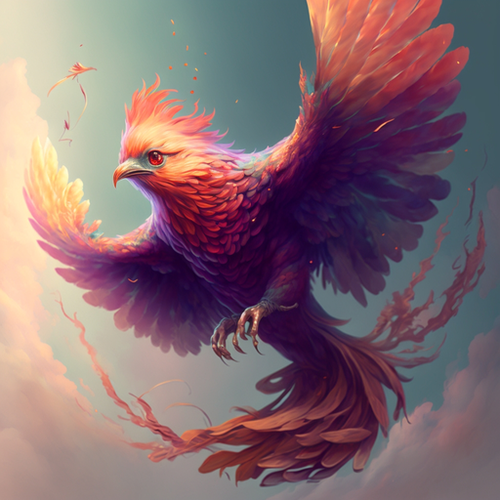}&%
        \includegraphics[width=0.7\aimvsipadaptertasksimgsize]{img/paper_imagery/vsipadapter_input_left.png}&%
        \includegraphics[width=0.7\aimvsipadaptertasksimgsize]{img/paper_imagery/vsipadapter_input_left.png}&%
        \includegraphics[width=0.7\aimvsipadaptertasksimgsize]{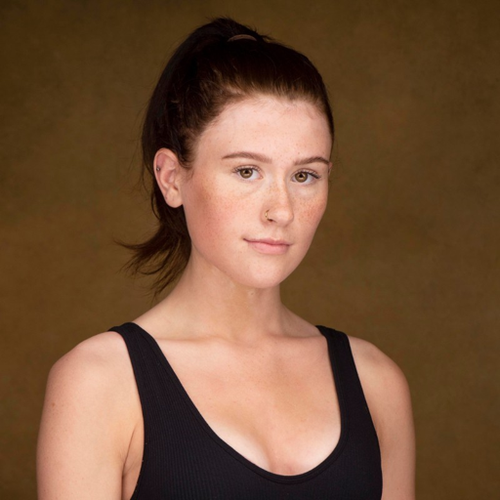}&%
        \includegraphics[width=0.7\aimvsipadaptertasksimgsize]{img/paper_imagery/vsipadapter_input_right.png}\\%
        &%
        \begin{tikzpicture}
            \draw[->]        (0,1)   -- (0,0) node [pos=0.75,above,fill=white,font=\footnotesize] {Recreate the image\vphantom{Ry}};
        \end{tikzpicture}&%
        \begin{tikzpicture}
        \draw[->]        (0,1)   -- (0,0) node [pos=0.75,above,fill=white,font=\footnotesize] {Transfer the style\vphantom{Ry}};
        \end{tikzpicture}&%
        \begin{tikzpicture}
        \draw[->]        (0,1)   -- (0,0) node [pos=0.75,above,fill=white,font=\footnotesize] {Extract the bird\vphantom{Ry}};
        \end{tikzpicture}&%
        \begin{tikzpicture}
            \draw[->]        (0,1)   -- (0,0) node [pos=0.75,above,fill=white,font=\footnotesize] {Copy composition\vphantom{Ry}};
        \end{tikzpicture}&%
        \begin{tikzpicture}
            \draw[->]        (0,1)   -- (0,0) node [pos=0.75,above,fill=white,font=\footnotesize] {Extract the face\vphantom{Ry}};
        \end{tikzpicture}\\%
        &%
        A flying phoenix\vphantom{Ry}
        &
        A bird house\vphantom{Ry}
        &
        A bird with a rock\vphantom{Ry}
        &
        A tabby cat\vphantom{Ry}
        &
        A girl with hat and scarf\vphantom{Ry}\\%
        \rotatebox{90}{\qquad\;~ IPAdapter+~\cite{IPAdapterPlusSD15}}&%
        \includegraphics[width=\aimvsipadaptertasksimgsize]{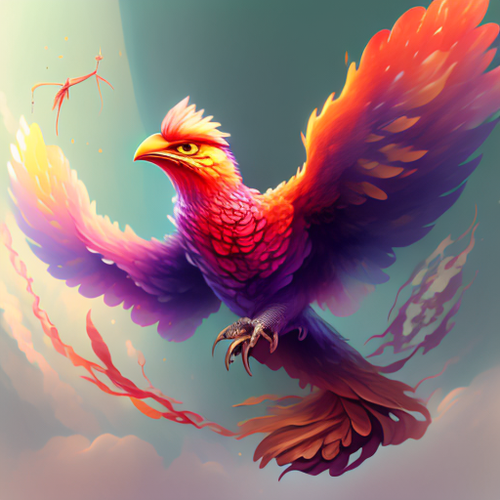}&%
        \includegraphics[width=\aimvsipadaptertasksimgsize]{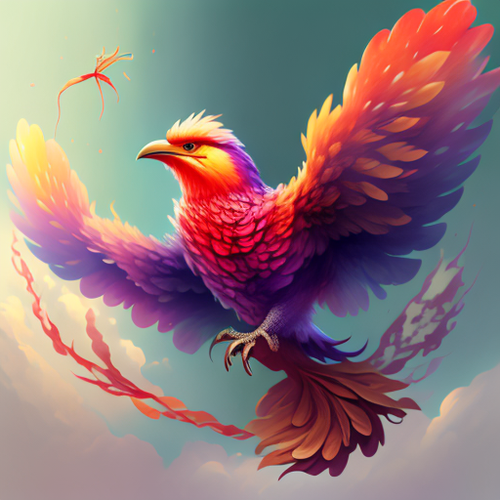}&%
        \includegraphics[width=\aimvsipadaptertasksimgsize]{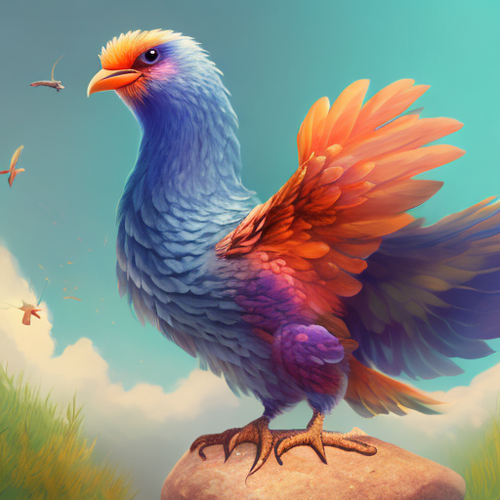}&%
        \includegraphics[width=\aimvsipadaptertasksimgsize]{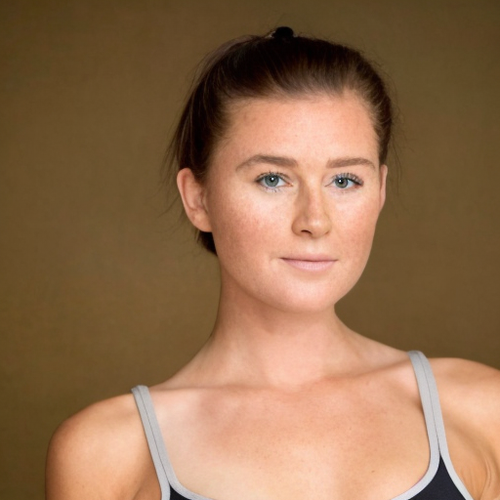}&%
        \includegraphics[width=\aimvsipadaptertasksimgsize]{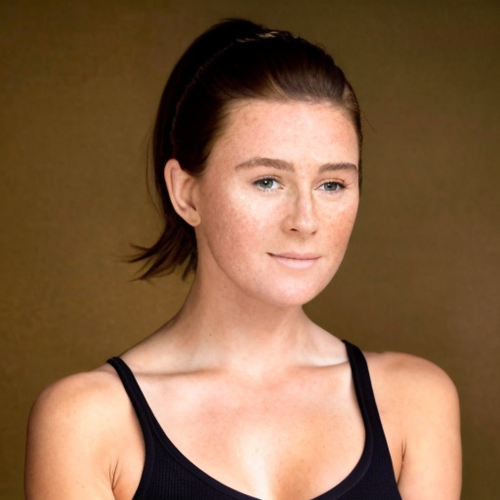}\\%
        \rotatebox{90}{\qquad\ InstructPix2Pix~\cite{brooks2023instructpix2pix}}&%
        \includegraphics[width=\aimvsipadaptertasksimgsize]{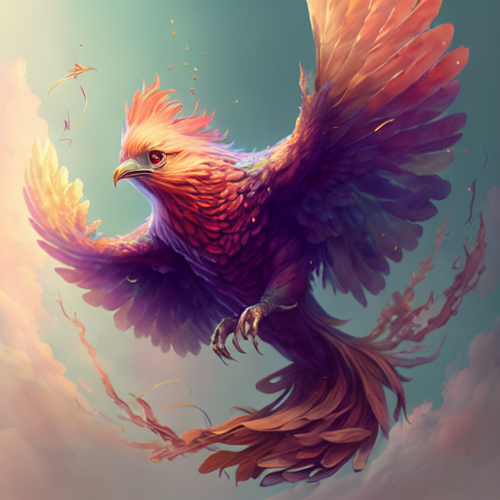}&%
        \includegraphics[width=\aimvsipadaptertasksimgsize]{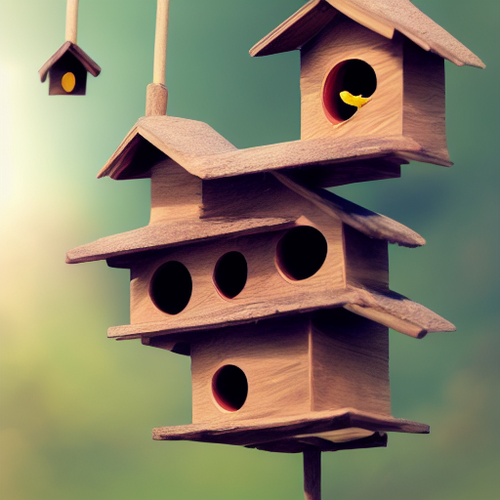}&%
        \includegraphics[width=\aimvsipadaptertasksimgsize]{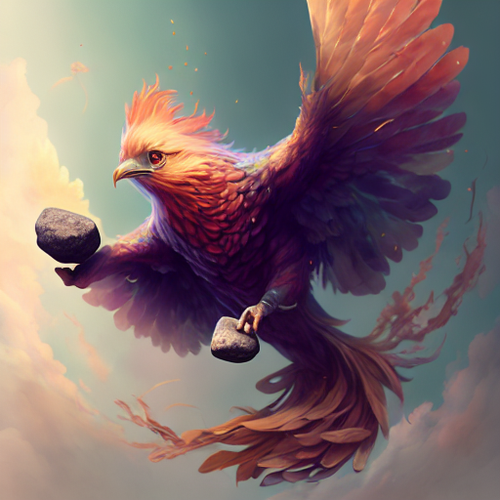}&%
        \includegraphics[width=\aimvsipadaptertasksimgsize]{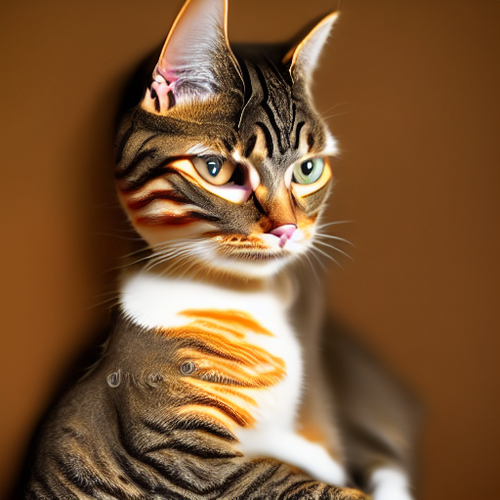}&%
        \includegraphics[width=\aimvsipadaptertasksimgsize]{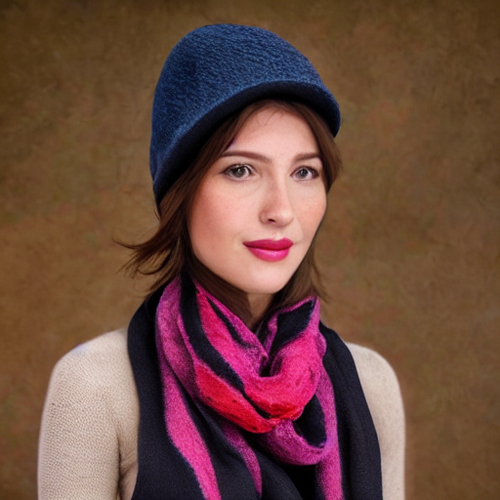}\\%
        \rotatebox{90}{\qquad\ IPAdapter-Instruct}&%
        \includegraphics[width=\aimvsipadaptertasksimgsize]{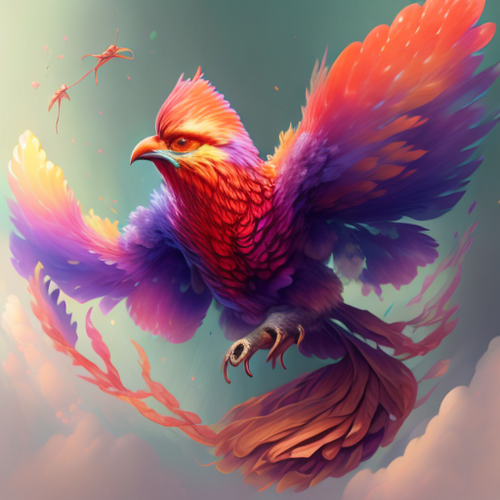}&%
        \includegraphics[width=\aimvsipadaptertasksimgsize]{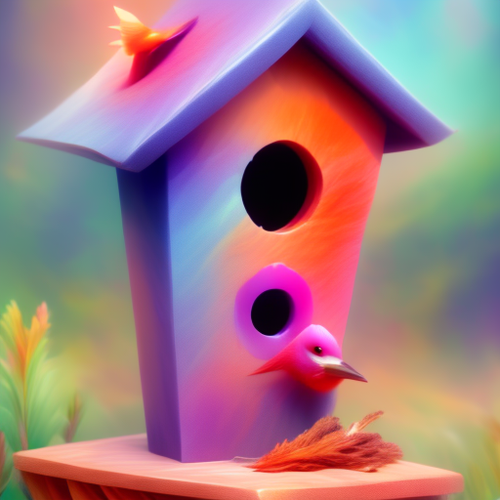}&%
        \includegraphics[width=\aimvsipadaptertasksimgsize]{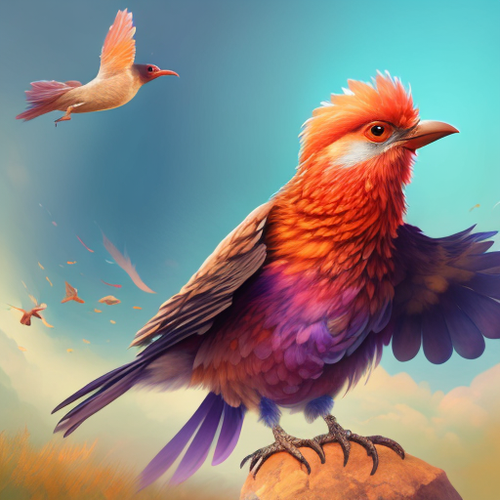}&%
        \includegraphics[width=\aimvsipadaptertasksimgsize]{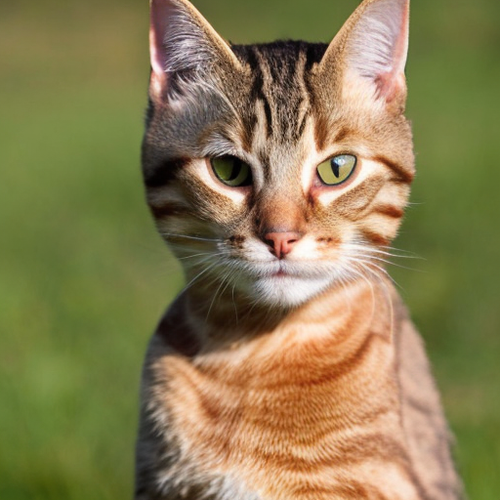}&%
        \includegraphics[width=\aimvsipadaptertasksimgsize]{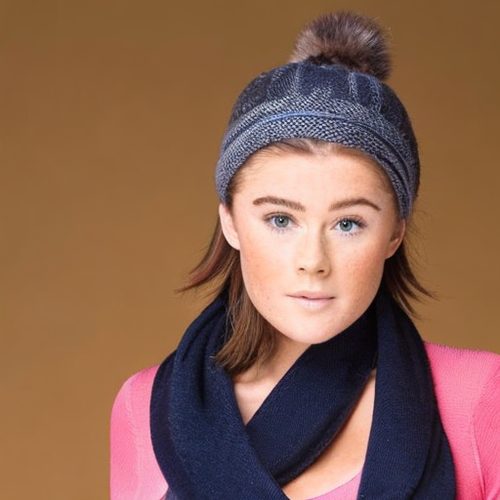}\\[3mm]%
        &%
        \includegraphics[width=0.7\aimvsipadaptertasksimgsize]{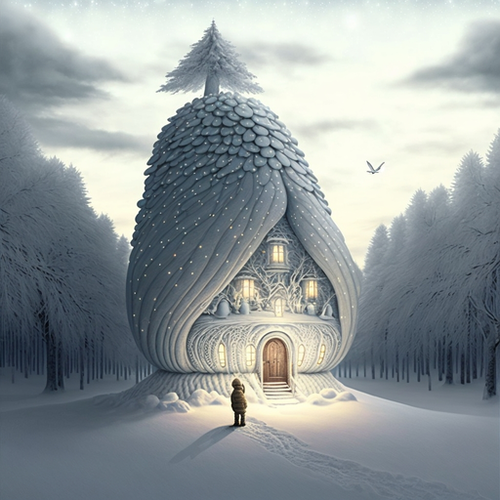}&%
        \includegraphics[width=0.7\aimvsipadaptertasksimgsize]{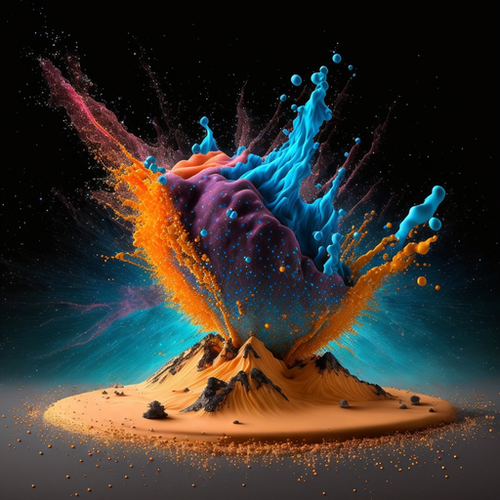}&%
        \includegraphics[width=0.7\aimvsipadaptertasksimgsize]{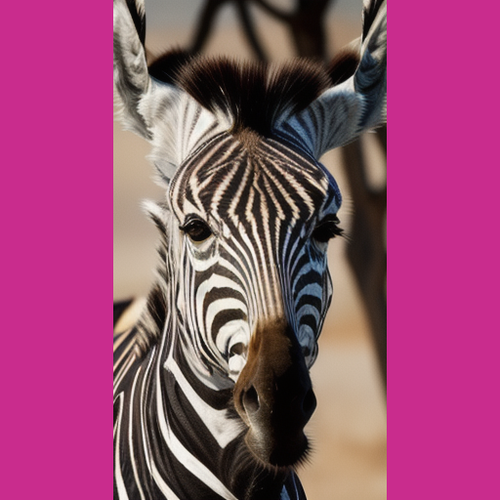}&%
        \includegraphics[width=0.7\aimvsipadaptertasksimgsize]{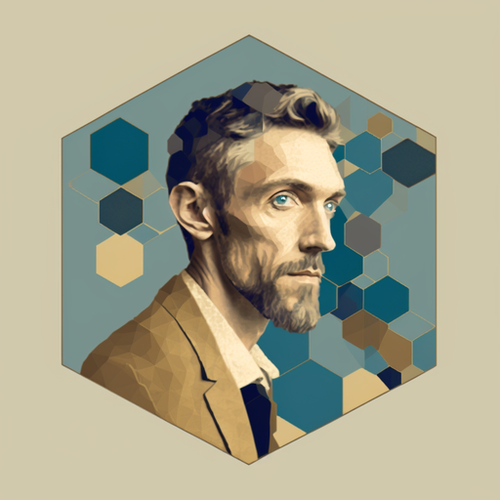}&%
        \includegraphics[width=0.7\aimvsipadaptertasksimgsize]{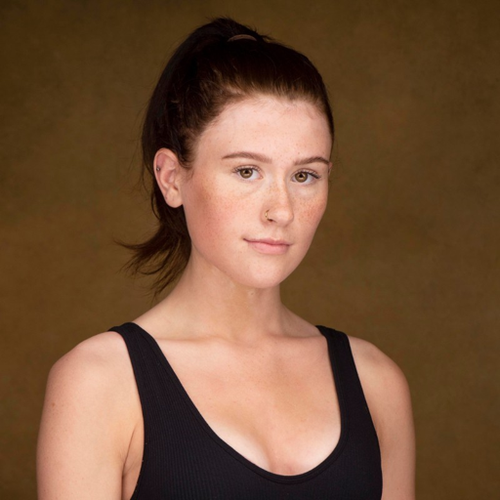}\\%
        &%
        \begin{tikzpicture}
            \draw[->]        (0,1)   -- (0,0) node [pos=0.75,above,fill=white,font=\footnotesize] {Recreate the image\vphantom{Ry}};
        \end{tikzpicture}&%
        \begin{tikzpicture}
        \draw[->]        (0,1)   -- (0,0) node [pos=0.75,above,fill=white,font=\footnotesize] {Transfer the style\vphantom{Ry}};
        \end{tikzpicture}&%
        \begin{tikzpicture}
        \draw[->]        (0,1)   -- (0,0) node [pos=0.75,above,fill=white,font=\footnotesize] {Extract the zebra\vphantom{Ry}};
        \end{tikzpicture}&%
        \begin{tikzpicture}
            \draw[->]        (0,1)   -- (0,0) node [pos=0.75,above,fill=white,font=\footnotesize] {Copy composition\vphantom{Ry}};
        \end{tikzpicture}&%
        \begin{tikzpicture}
            \draw[->]        (0,1)   -- (0,0) node [pos=0.75,above,fill=white,font=\footnotesize] {Extract the face\vphantom{Ry}};
        \end{tikzpicture}\\%
        &%
        A round house in winter
        &
        A chest full of rich clothing
        &
        A palm tree like a zebra
        &
        A dog's face
        &
        A woman holding a sword\\%
        \rotatebox{90}{\qquad\;~ IPAdapter+~\cite{IPAdapterPlusSD15}}&%
        \includegraphics[width=\aimvsipadaptertasksimgsize]{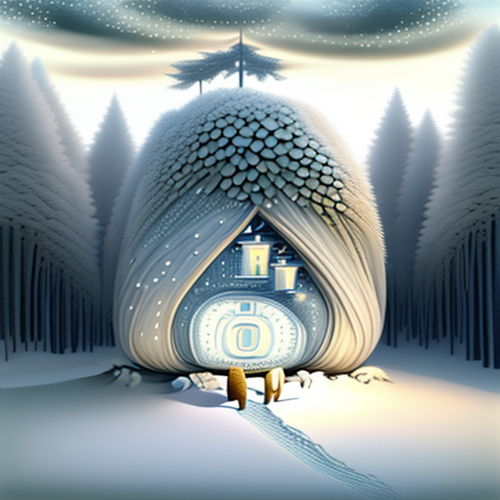}&%
        \includegraphics[width=\aimvsipadaptertasksimgsize]{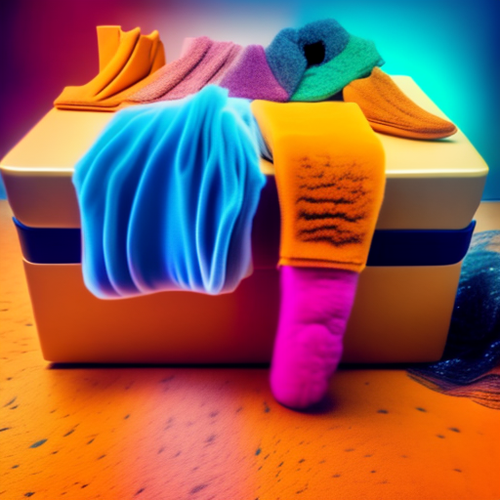}&%
        \includegraphics[width=\aimvsipadaptertasksimgsize]{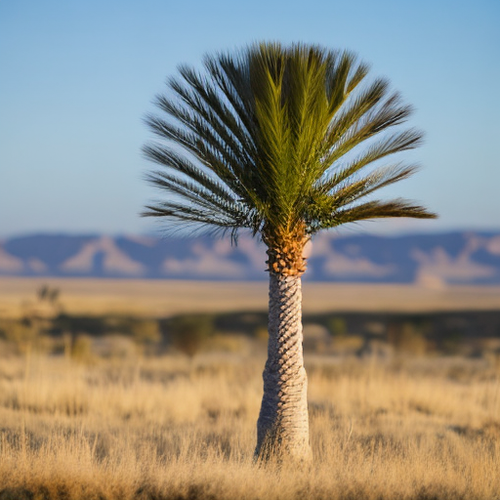}&%
        \includegraphics[width=\aimvsipadaptertasksimgsize]{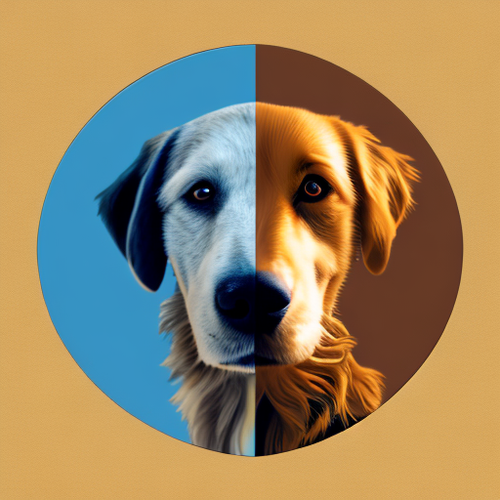}&%
        \includegraphics[width=\aimvsipadaptertasksimgsize]{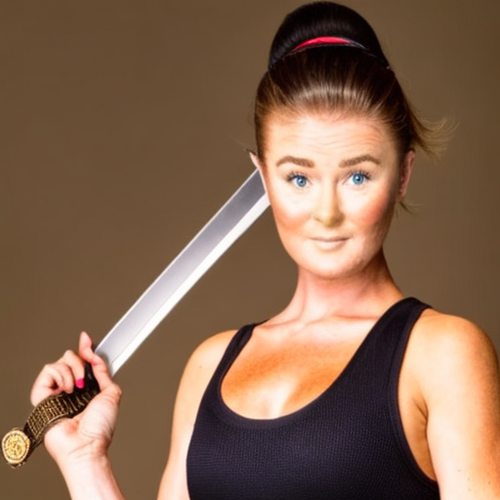}\\%
        \rotatebox{90}{\qquad\ InstructPix2Pix~\cite{brooks2023instructpix2pix}}&%
        \includegraphics[width=\aimvsipadaptertasksimgsize]{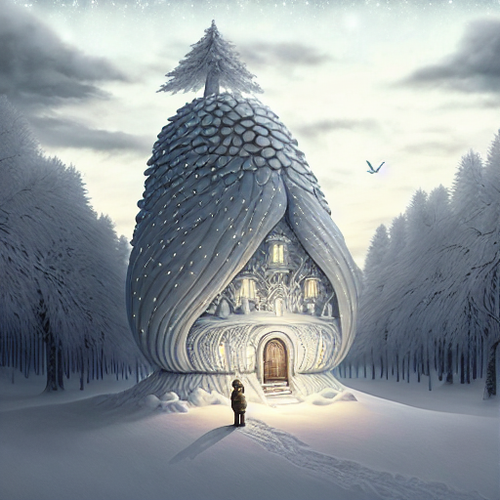}&%
        \includegraphics[width=\aimvsipadaptertasksimgsize]{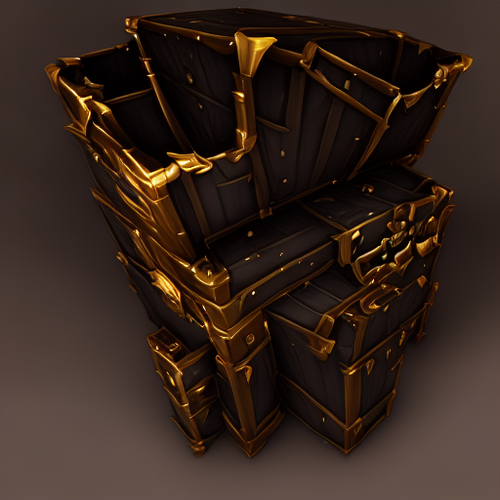}&%
        \includegraphics[width=\aimvsipadaptertasksimgsize]{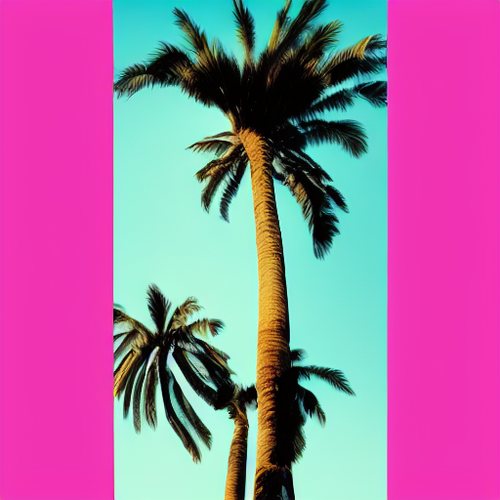}&%
        \includegraphics[width=\aimvsipadaptertasksimgsize]{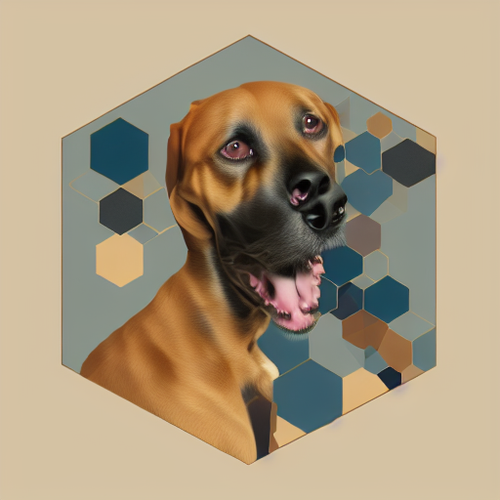}&%
        \includegraphics[width=\aimvsipadaptertasksimgsize]{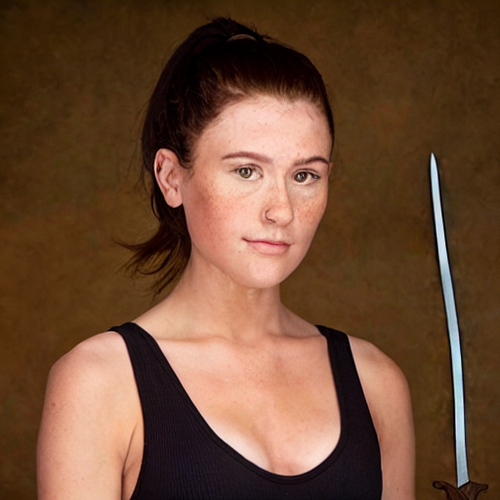}\\%
        \rotatebox{90}{\qquad\ IPAdapter-Instruct}&%
        \includegraphics[width=\aimvsipadaptertasksimgsize]{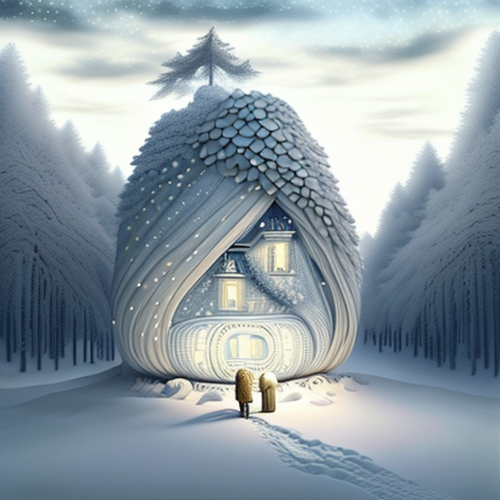}&%
        \includegraphics[width=\aimvsipadaptertasksimgsize]{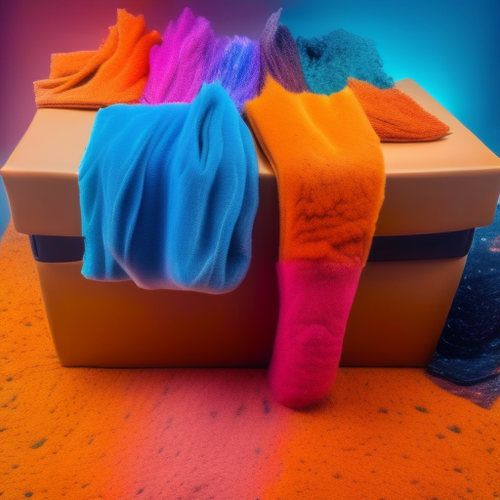}&%
        \includegraphics[width=\aimvsipadaptertasksimgsize]{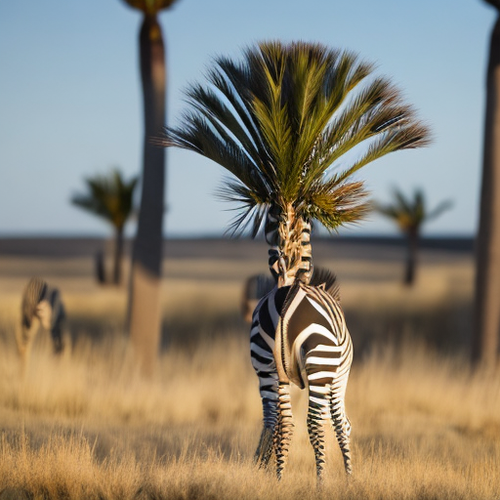}&%
        \includegraphics[width=\aimvsipadaptertasksimgsize]{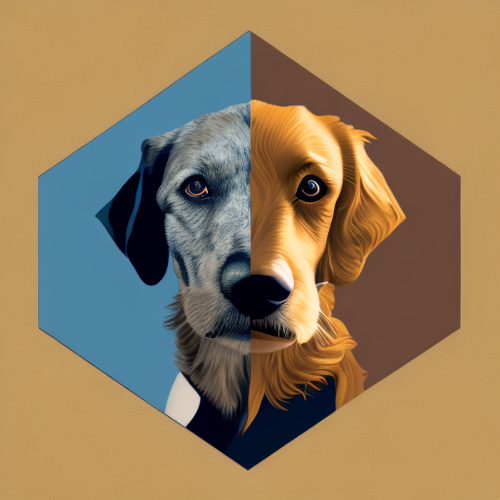}&%
        \includegraphics[width=\aimvsipadaptertasksimgsize]{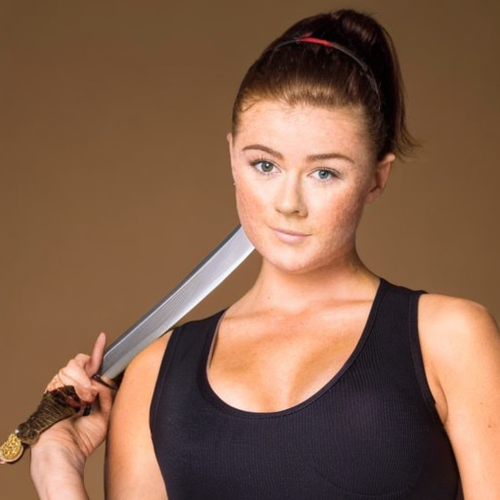}%
    \end{tabular}
}%

\end{center}
\captionof{figure}{Qualitative examples of the proposed approach compared to IPAdapter+~\cite{IPAdapter,IPAdapterPlusSD15} and InstructPix2Pix~\cite{brooks2023instructpix2pix}.
IPAdapter fails whenever the user intent is not clear, while InstructPix2Pix fails whenever the output should differ significantly from the input.
Please refer to \cref{sec:qual-results} for a detailed discussion.
}\label{fig:vs-ipadapter-single-image}
\end{minipage}

\clearpage

\begin{figure}[t]
    \begin{subfigure}[b]{0.48\linewidth}
    \centering
    \begin{tabular}{cc}\scriptsize%
        Style&\scriptsize%
        Scribble\\%
        \includegraphics[width=0.35\linewidth]{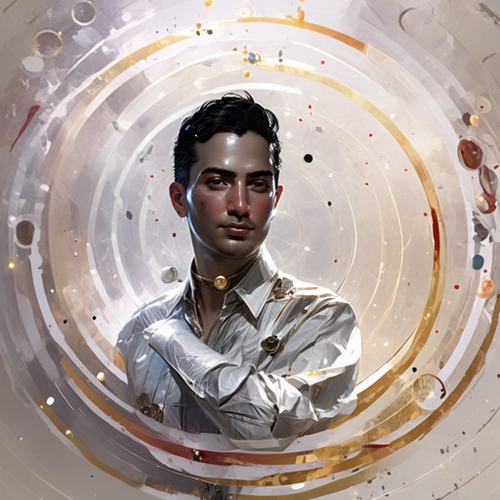}&%
        \includegraphics[width=0.35\linewidth]{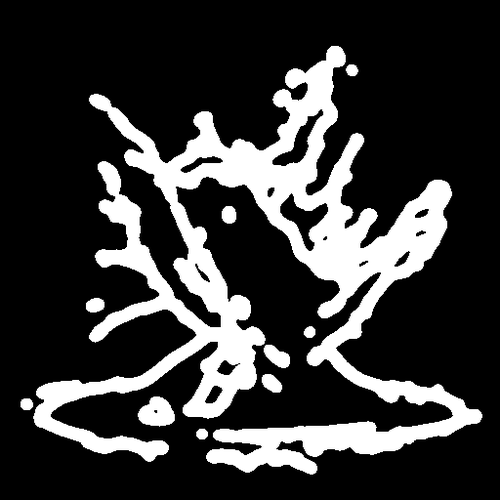}\\[2mm]\scriptsize%
        Oyster with a pearl&\scriptsize%
        Mountain explosion\\%
        \includegraphics[width=0.45\linewidth]{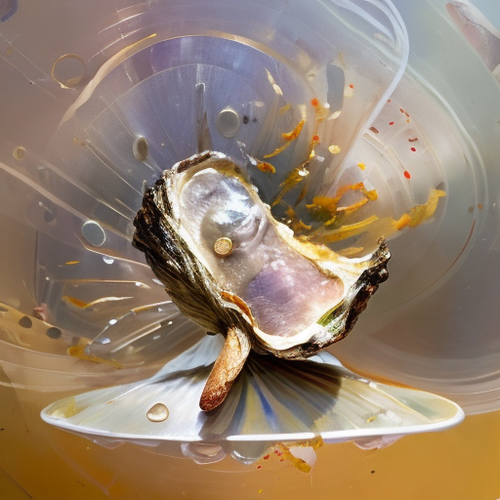}&%
        \includegraphics[width=0.45\linewidth]{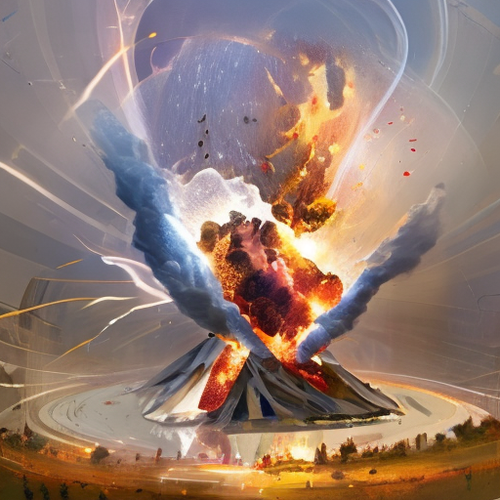}%
    \end{tabular}
    \caption{Combining IPAdapter-Instruct with a style instruction and a scribble ControlNet~\cite{ControlNet,ControlNetScribble} to only control style without bleeding other concepts from the condition works extremely well.}
    \end{subfigure}
    \hfill
    \begin{subfigure}[b]{0.48\linewidth}
    \centering
    \begin{tabular}{cc}\scriptsize%
        Face&\scriptsize%
        Pose\\%
        \includegraphics[width=0.35\linewidth]{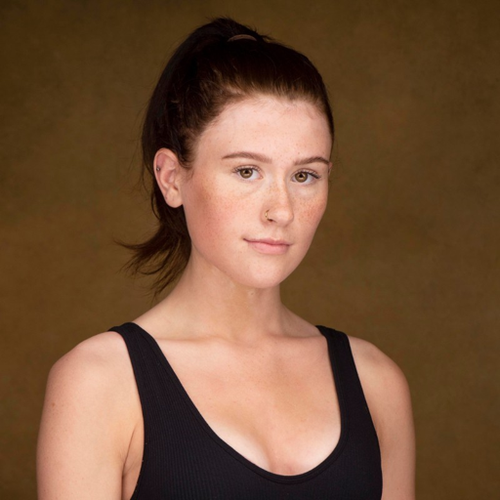}&%
        \includegraphics[width=0.35\linewidth]{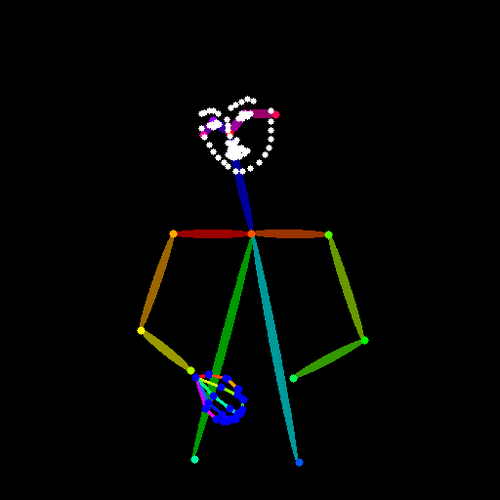}\\[2mm]\scriptsize%
        Female Chef&\scriptsize%
        Female Firefighter\\%
        \includegraphics[width=0.45\linewidth]{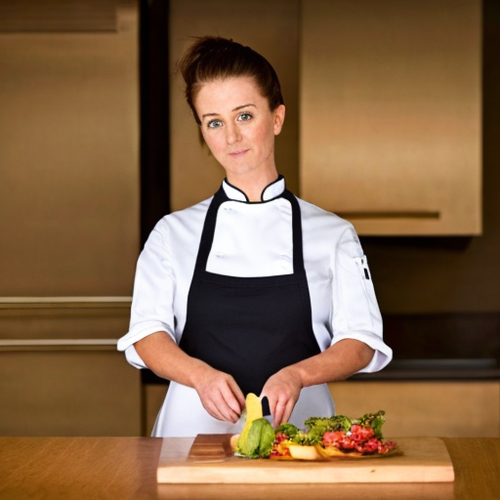}&%
        \includegraphics[width=0.45\linewidth]{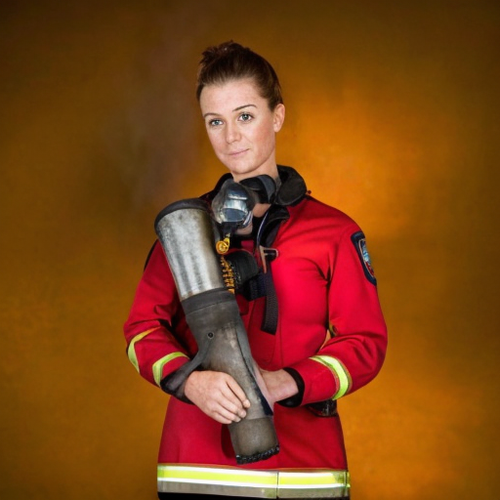}%
    \end{tabular}
    \caption{Similarly, we can use IPAdapter-Instruct with a face extraction instruction in combination with a pose ControlNet~\cite{ControlNet,ControlNetPose} for fine-grained control over the final pose.}
    \end{subfigure}
\caption{Examples of the interoperability of IPAdapter-Instruct with ControlNets for the underlying image diffusion model.}\label{fig:controlnet-compatibility}
\end{figure}

Finally, similar to IPAdapter but contrary to InstructPix2Pix, we retain compatibility with ControlNet and LoRA models, as seen in \cref{fig:controlnet-compatibility}.
Our model succesfully conditions the generation process as before, while the ControlNets offers pixel-precise guidance.

\section{Conclusion, Limitations, and Future Work}\label{sec:conclusion}%

In this work, we have introduced IPAdapter-Instruct to disambiguate user intent when conditioning image diffusion models on input images: by introducing an instruction prompt that specifies user intent, this joint model can be trained efficiently without losing performance.
This compacts multiple adapters into a single prompt and image combination, while retaining the benefits from keeping the base diffusion model intact, such as remaining compatible with its LoRAs.

We found the main limitation to be the creation of the training datasets: it is time-consuming and strongly restricted by source data availability, but has a significant impact on the task performances.
They also clearly impose biases on the conditioning model: style transfer is biased towards MidJourney~\cite{JourneyDB}, face extraction works best on real photos, and --- most strikingly --- object extraction benefits significantly from colored padding of the inputs.

We find that our model struggles most with the composition task: even when structure is preserved, it tends to bleed style.
This is not unexpected: the composition task (as pixel-precise guidance) is arguably better suited for ControlNet or InstructPix2Pix.
We envision future work to combine both InstructPix2Pix and IPAdapterInstruct into a single instruction-conditioned model.

\bibliographystyle{splncs04}
\bibliography{main,add}

\end{document}